\documentclass[journal]{IEEEtran}

\ifCLASSINFOpdf
\else
\fi

\usepackage{graphicx}
\usepackage{amssymb}
\usepackage{epstopdf}
\usepackage{subfigure}
\usepackage{algorithm}
\usepackage{algorithmic}
\usepackage{amsmath}

\newtheorem{definition}{Definition}

\begin{document}

\title{Discretization of Parametrizable Signal Manifolds}

\author{Elif Vural and Pascal Frossard
\thanks{E. Vural and P. Frossard are with Ecole Polytechnique F\'{e}d\'{e}rale de Lausanne (EPFL), Signal Processing Laboratory - LTS4, CH-1015 Lausanne, Switzerland. email: elif.vural@epfl.ch, pascal.frossard@epfl.ch.}
\thanks{This work has been partly funded by the Swiss National Science Foundation under Grant  $200021\_120060$.}}


\maketitle

\begin{abstract}
Transformation-invariant analysis of signals often requires the computation of the distance from a test pattern to a transformation manifold. In particular, the estimation of the distances between a transformed query signal and several transformation manifolds representing different classes provides essential information for the classification of the signal. In many applications the computation of the exact distance to the manifold is costly, whereas an efficient practical solution is the approximation of the manifold distance with the aid of a manifold grid.  In this paper, we consider a setting with transformation manifolds of known parameterization. We first present an algorithm for the selection of samples from a single manifold that permits to minimize the average error in the manifold distance estimation. Then we propose a method for the joint discretization of multiple manifolds that represent different signal classes, where we optimize the transformation-invariant classification accuracy yielded by the discrete manifold representation. Experimental results show that sampling each manifold individually by minimizing the manifold distance estimation error outperforms baseline sampling solutions with respect to registration and classification accuracy. Performing an additional joint optimization on all samples improves the classification performance further. Moreover, given a fixed total number of samples to be selected from all manifolds, an asymmetric distribution of samples to different manifolds depending on their geometric structures may also increase the classification accuracy in comparison with the equal distribution of samples.

\end{abstract}


\IEEEpeerreviewmaketitle

\section{Introduction}
\label{sec:Intro}

\IEEEPARstart{C}{ompared} to traditional signal processing techniques where the treatment of signals is performed in the high-dimensional signal space, the characterization of signal sets via low-dimensional manifold models have several advantages. Particularly in image processing, manifold models have been seen to provide more compact and efficient signal representations as well as assisting the analysis of data. Manifold models for signals have recently been studied in several research areas such as dimensionality reduction \cite{266187}, \cite{donoho03hessian}, image registration \cite{Fitzgibbon03}, \cite{10.1109/TPAMI.2008.156} and transformation-invariant pattern classification \cite{DBLP:journals/tmm/VasconcelosL05}.

A signal manifold in a high-dimensional signal space is a set of signals that can be mapped to a lower-dimensional parameter space. In \cite{1486387} Peyr\'{e} et al. demonstrate several examples of parametrizable signals using local patch manifolds, and examine inverse problem regularizations with manifold models and applications to image inpainting. Similarly, Wakin et al. study image appearance manifolds (IAM) in \cite{4754117}, which are image sets defined by a few generating articulation parameters. In this work, we consider transformation manifolds that are defined as signal sets which can be described by parametrizable transformations of a reference signal model. An instance of transformation manifold can be a pattern transformation manifold, which refers to the family of patterns obtained by applying a certain set of geometric transformations to a reference pattern; or the observation manifold of an object, which consists of its images captured under varying viewpoints. Among several manifold models, transformation manifolds are of specific interest as they provide a basis for performing transformation-invariant analysis of signals that have been exposed to geometric transformations. For instance, it is possible to classify a transformed query signal by computing the distance between the signal and each of the transformation manifolds that represent different classes. The class label of the signal is then estimated to be the same as the class label of the manifold that has the smallest distance to the signal.

The computation of the manifold distance is in general a demanding task, mainly due to the variety and complexity of the involved models. There have been several research efforts addressing the manifold distance computation problem. The tangent distance provides an estimation of the manifold distance through local linearity assumptions \cite{668381}. However, this approach is likely to suffer from local minima, which is improved by the investigation of the tangent distance in a multiresolution manner \cite{DBLP:journals/tmm/VasconcelosL05}. While these are generic methods applicable to a large set of manifold-modeled signals, there are also some works proposing transformation estimation solutions for particular types of manifolds, such as those generated by 2D pattern transformations \cite{10.1109/TPAMI.2008.156}, \cite{bb25414}.

Although there have been many attempts, the general problem of manifold distance estimation still retains its challenges as it usually involves high computational complexity. In order to estimate the manifold distance, it is a practical solution to represent the manifold by a finite grid of manifold samples, where the distance between a space point and its projection onto the manifold is approximated by the distance between the point and the nearest manifold sample. The usage of such a grid improves the distance estimation complexity immensely, possibly at the price of a lower distance accuracy. 

The choice of the manifold grid has considerable influence on the accuracy of the distance estimation. In this work, we study the distance-based discretization of signal manifolds of known parameterization. We build on our previous work, where we propose a manifold discretization algorithm that minimizes the manifold distance estimation error stemming from the representation of the manifold by finitely many grid points \cite{ManifoldDisc}. Our discretization method bears some resemblance to the LBG vector quantization algorithm \cite{1094577} due to the alternating steps of optimization that compute the representative samples for a given space partition, and respectively the partition for a given set of samples. However, the proposed method differs essentially from the LBG algorithm, since it targets the minimization of the manifold distance with samples positioned on the manifold and does not have a signal approximation objective. Noting the dependency between the registration and classification efficiencies, we extend this sampling solution to a setting with several transformation manifolds representing different class models, where we optimize all manifold samples in a joint manner such that the relative geometries of different manifolds are taken into account. The experimental results show that our discretization approach is significantly better than basic discretizations such as random grids or regular grids in the parameter domain in terms of registration and classification accuracy. The classification performance is further improved when the relative properties of the manifolds are also considered in addition to their individual properties in the sampling.

Meanwhile, some state-of-the-art solutions to surface discretization are as follows. It is typical to achieve the sampling in a straightforward way by generating a grid regular in the parameter domain. However, especially when the number of manifold samples is limited, a regular discretization in the parameter space is not guaranteed to offer a good performance. Structured grid generation has been well-studied especially for analytical two-dimensional surfaces in $ \mathbb R^3$, mostly for the purpose of obtaining finite-difference solutions to partial differential equations \cite{BasicStructuredGridGen}. On the other hand, it is also possible to find sampling solutions for surfaces represented in non-analytical forms such as meshes \cite{1133794}. Even though some of these sampling methods may in principle be generalized for arbitrary dimensional manifolds, the targeted applications must be taken into account in grid generation. 

The paper is organized as follows: In Section \ref{sec:Notations}, we formulate the utilization of transformation manifolds in signal classification and set our notation. In Section \ref{sec:distanceBasedManDisc} we overview the discretization of parametric signal manifolds based on distance estimation, whereas in Section \ref{sec:classBasedManDisc} we propose an extension of the registration-based sampling solution for classification. We present the experimental results obtained on some transformation manifolds in Section \ref{sec:ExpRes}, and conclude in Section \ref{sec:Conclusion}.

\section{Manifold Models in Transformation-Invariant Signal Analysis}
\label{sec:Notations}

The analysis of transformed signals in registration or classification settings mostly consists in estimating their distance or projection to transformation manifolds. Let $\mathcal{M} \subset \mathbb R^n $ denote a set of signals conforming to a manifold model defined over the parameter domain $\Lambda \subset  \mathbb R^d $. Then, $\mathcal{M}$ is given by $\mathcal{M} = \{U(\lambda),  \lambda \in \Lambda  \}$, where $U$ denotes the mapping from the $d$-dimensional parameter domain $\Lambda$ to the set of $n$-dimensional signals. As an example, consider a reference visual pattern, and the family of patterns generated by applying arbitrary rotations and scale changes to the reference pattern. Then, every combination of rotation $\psi$ and scale $s$ defines a parameter vector $\lambda=(\psi,s)$, and the set of constructed patterns lies on a pattern transformation manifold $\mathcal{M} = \{U(\lambda),   \lambda = (\psi, s)  \}$ of intrinsic dimension $d=2$. Image appearance manifolds \cite{4754117} or image parameter articulation manifolds \cite{2813257} constitute other examples of the manifolds we study. Although we focus on transformation manifolds in this work, we maintain a generic formulation that it is applicable to parametric signal manifolds in general. We also note that parametrizable signal manifolds are not restricted to image manifolds, which could find examples within acoustic and seismic signals for instance \cite{1486387}, \cite{AnalysisSeismicWave}.

Let $x \in \mathbb R^n$ be a signal belonging to the space of $n$-dimensional digital signals. We denote the distance of $x$ to the manifold $\mathcal{M}$ as $D(x,\mathcal{M})= \min_{\lambda  \in \Lambda} \{ d(x,U(\lambda)) \}$, where the distance function $d(x,y)$ is taken as the $\ell_2$-distance $\| x-y\|$ throughout this work. Then, a projection $y$ of $x$ onto the manifold is a manifold point with minimum distance to $x$, i.e., $y = U(\lambda ^{\ast} )$, for some $\lambda ^{\ast}$ such that $\lambda ^{\ast} = \arg \min_{\lambda  \in \Lambda} \{ d(x,U(\lambda)) \}$. The projection of a signal onto a transformation manifold is typically associated with the registration of  the signal with respect to a transformation model. Given a signal $x$ with an unknown geometric transformation and the related transformation manifold $\mathcal{M}$, the estimation of the transformation that best represents $x$ corresponds to finding the point of $\mathcal{M}$ with smallest distance to $x$.

Now let us consider a setting with multiple signal manifolds, each of which represents the transformation model of a different class. Let $\mathcal{M}_1, \mathcal{M}_2, \cdots , \mathcal{M}_M \subset  \mathbb R^n$ be smooth\footnote{In \cite{4754117} it is shown that differentiability of a manifold fails in the cases where the generating model involves sharp edges, however, approximate differentiable representations of such manifolds can be obtained by their regularization.} manifolds of $M$ different signal classes defined on the parameter domains $ \Lambda_1,  \Lambda_2, \cdots,  \Lambda_M$  by the mappings $U_1, U_2, \cdots , U_M$, where $U_m: \Lambda_m \rightarrow \mathcal{M}_m$, $m=1,\cdots,M$. We assume that all $\{ \mathcal{M}_m \}$ are submanifolds of the same Euclidean space $\mathbb R^n$, however, the parameter domains $\{ \Lambda_m \}$ may be subsets of different spaces. Let $x\in \mathbb R^n$ be a query signal of unknown class that has undergone an unknown transformation. Then, provided that the transformation manifolds $\{ \mathcal{M}_m \}$ constitute sufficiently accurate models for the representation of the involved signal classes, $x$ belongs to the same class as the manifold with minimum distance to it, i.e., the class label $l(x)$ of $x$ is given by

\begin{equation}
l(x)=\arg \min_m D(x,\mathcal{M}_m), \, \, m=1,\cdots, M.
\end{equation}

Now let us consider only two manifolds $\mathcal{M}_m$ and $\mathcal{M}_r$ representing two different class models. In order to represent the sets of signals that belong to the same class as each of these two manifolds, we define the half-spaces\footnote{Although in the standard definition, the boundary surface determining a half-space is an affine hyperplane, here we generalize the term to include the case where the boundary is a hypersurface.} $\mathcal{H}_{mr}$ and $\mathcal {H}_{rm}$ as follows: We denote by $\mathcal{H}_{mr}$ the set of points whose distance to $\mathcal{M}_m$ is smaller than their distance to $\mathcal{M}_r$ (notice that $\mathcal{H}_{mr}\neq \mathcal {H}_{rm}$ with our notation):
\begin{definition}
\begin{equation}
\label{eq:defnHmr}
\mathcal{H}_{mr}=\overline{  \{   x \in \mathbb R^n  :  D(x,\mathcal{M}_m) < D(x,\mathcal{M}_r)  \} }.
\end{equation}
\end{definition}
The notation $\overline{S }$ in (\ref{eq:defnHmr}) denotes the closure of the set $S$. Note that $\mathcal{H}_{mr}$ is defined in this way considering the degenerate cases that may be caused by manifold intersections. 
We then define the decision surface $\mathcal{B}_{mr}$ as the boundary of the half-space $\mathcal{H}_{mr}$,

\begin{definition}
\label{def:EquidistanceSurface}
\begin{equation}
\begin{split}
\mathcal{B}_{mr}=\partial \mathcal{H}_{mr}. 
\end{split}
\end{equation}
\end{definition}
$\mathcal{B}_{mr}$ is a combination of hypersurfaces, i.e.~a union of $(n-1)$-dimensional manifolds in $\mathbb R^n$.

\begin{figure}
 \centering
  \includegraphics[scale=0.7]{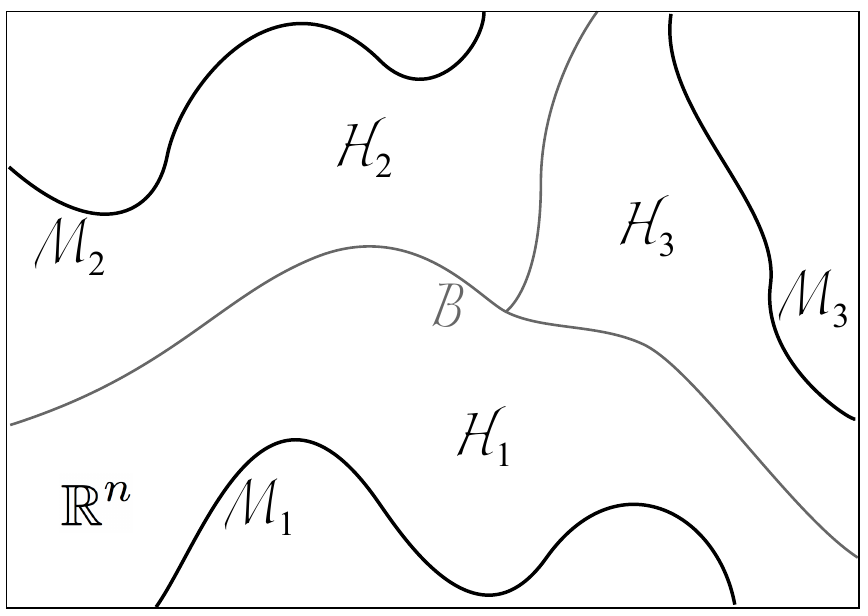}
  \caption{Illustration of transformation-invariant signal classification via transformation manifolds: $\mathcal{M}_1$, $\mathcal{M}_2$, $\mathcal{M}_3$ are three manifolds representing different classes; $\mathcal{H}_1$, $\mathcal{H}_2$, and $\mathcal{H}_3$ are their respective approximation regions; and $\mathcal{B}$ is the combined decision surface.}
  \label{fig:illustrationManifolds1}
\end{figure}

If we consider $M$ class representative manifolds instead of two, the set of space signals belonging to the class represented by $\mathcal{M}_m$ are given by the approximation region $\mathcal{H}_m \subset  \mathbb R^n$ of $\mathcal{M}_m$:
\begin{definition}
\begin{equation}
\mathcal{H}_m= \bigcap_{r \in \{1,\cdots,M\} \setminus \{ m\} } \mathcal{H}_{mr}\, . 
\end{equation}
\end{definition}
Finally, in order to adapt the decision surface $\mathcal{B}_{mr}$ determined by two manifolds to the case of multiple manifolds, we define the combined decision surface $\mathcal{B}$,
\begin{definition}
\begin{equation}
\mathcal{B}=\bigcup_{m=1}^M \partial   \mathcal{H}_m,
\end{equation}
\end{definition}
where the notation $\partial S$ denotes the boundary of the set $S$. The combined decision surface $\mathcal{B}$ is a collection of subsets of all decision surfaces $\mathcal{B}_{mr}$, i.e., $\mathcal{B} \subset \bigcup_{m \neq r} \mathcal{B}_{mr} $. It forms a boundary between different portions of the space, where each of these portions belongs to a different class as imposed by the manifolds $\mathcal{M}_1, \mathcal{M}_2, \cdots , \mathcal{M}_M $. An illustration of class representative transformation manifolds, the approximation regions $\mathcal{H}_m$ of manifolds, and the decision surface $\mathcal{B}$ is given in Fig.~\ref{fig:illustrationManifolds1}.

Now, having set our notation, we remark the following. The registration problem consists in obtaining the projection of a signal $x$ onto a manifold $\mathcal{M}$, and the classification of the signal $x$ corresponds to the determination of the approximation region $\mathcal{H}_m$ it lies in. The exact knowledge of the manifolds determines the projection and class label of a signal perfectly. 
Although a discrete representation is less accurate, it reduces significantly the complexities of the registration and classification problems, and the requirement for storage space. The accuracy of the discretization depends on its capability in approximating the manifold distance. In Sections \ref{sec:distanceBasedManDisc} and \ref{sec:classBasedManDisc}, the registration and classification accuracies of discretizations are formulated and sampling solutions are proposed.

\section{Manifold Discretization for Minimal Distance Estimation Error}
\label{sec:distanceBasedManDisc}

In this section, we describe a solution for manifold discretization such that the manifold distance estimation error caused by representing the manifold with a finite number of samples is minimized. Note that the main purpose of the sampling scheme proposed here is the accurate estimation of the manifold distance for registration applications rather than the approximation of the manifold, which are not necessarily equivalent. We build on \cite{ManifoldDisc} and present an iterative approach for the optimization of the manifold samples.

The discretization of a manifold $\mathcal{M}$ consists in selecting a predetermined number $N$ of manifold points, i.e., a sample set $\mathcal{S}= \{S_i\}=   \{ U(\lambda_i) \}  \subset \mathcal{M}$, $i = 1, \cdots, N $ for some  $\{ \lambda_1, \cdots, \lambda_N \} \subset \Lambda $.

We would like to select a set of samples that minimizes the total manifold distance estimation error $E$ over $R$, where $R$ is a bounded and piecewise connected region in the space $\mathbb R^n$. We consider $R$ to be a region of interest which depends on the application. For instance, in the case of images one can define $R$ as a hyperrectangular region $ R=\{ x\in \mathbb R^n : a_i \leq x_i \leq b_i \}$ with the typical values of 0 and 255 for the parameters $a_i$ and $b_i$. We define the error $E$ by
\begin{equation}  E=\int_R ( D^2(x,\mathcal{S}) - D^2(x,\mathcal{M}) ) \,dx,  \label{eq:E_total}  \end{equation}
where $D(x,\mathcal{S})=\min_{ i=1,2, \cdots  ,N}\{ d(x,S_i) \} $ denotes the distance between $x$ and the sample set $\mathcal{S}$. The formulation of the error in terms of the squared distances is due to the ease of analytical manipulation.

For a given sample set, one can partition $R$ into $N$ regions as $R=\bigcup_{i=1}^N R_i$, where each $R_i$ is a region consisting of points with smallest $\ell_2$-distance to $S_i$ among all samples, i.e., $R_i=\{x\in R: d(x,S_i) \leq d(x,S_j), \forall j \in \{1, \cdots, N  \}  \}$. Hence, the total manifold distance estimation error becomes
\begin{equation} E=\sum_{i=1}^N E_i = \sum_{i=1}^N \int_{R_i}  ( d^2(x,S_i) - D^2(x,\mathcal{M})  ) \,dx. 
\end{equation}

In order to minimize the error $E$, we follow an iterative optimization procedure. In each iteration of the algorithm, we first determine the partition regions corresponding to the samples, and then optimize each sample individually such that the error $E_i$ in the regarding region is minimized. Once the partition is determined, the minimization of the manifold distance estimation error $E_i$ within a specific region $R_i$ is achieved as follows. The error term $E_i$ can be rearranged as
\[ E_i= \int_{R_i}  d^2(x,S_i) \,dx - \int_{R_i} D^2(x,\mathcal{M}) \,dx,   \]

\noindent
where the second integration depends only on $R_i$, and is constant with respect to $S_i$. Therefore, $E_i$ is given by
\begin{eqnarray*} 
E_i&=& \int_{R_i}  d^2(x,S_i) \,dx + c_i\\
      &=& \int_{R_i} x^\mathrm T x \,dx  - 2 S_i^\mathrm T  \int_{R_i} x \,dx  + V_i S_i^\mathrm T S_i + c_i ,\\
\end{eqnarray*}
where $c_i$ is a constant independent of $S_i$, and $V_i= \int_{R_i}\,dx$ is the volume of the region $R_i$. Denoting the centroid of $R_i$ by $G_i= 
({\int_{R_i}  x \,dx  }) / ({\int_{R_i}\,dx})$, we get
\begin{eqnarray*}
E_i&=& \int_{R_i} x^\mathrm T x \,dx   +V_i( - 2 S_i^\mathrm T G_i +S_i^\mathrm T S_i) + c_i\\
      &=&V_i( S_i^\mathrm T S_i - 2 S_i^\mathrm T G_i) + c_i',\\ 
 \end{eqnarray*}
where we express the sum of the terms independent of $S_i$ by $c_i'$. As $E_i$ differs from $\| S_i - G_i \|^2$ only up to a positive multiplicative factor and an additive term constant with respect to $S_i$, one can equivalently minimize
\begin{equation} \varepsilon_i= \| S_i - G_i \|^2 \label{eq:Ei_final_norm}  \end{equation}
at each iteration of the algorithm. This actually means that $S_i$ should be selected as the manifold point closest to the centroid of the region $R_i$.

\begin{figure}
 \centering
  \includegraphics[scale=0.54]{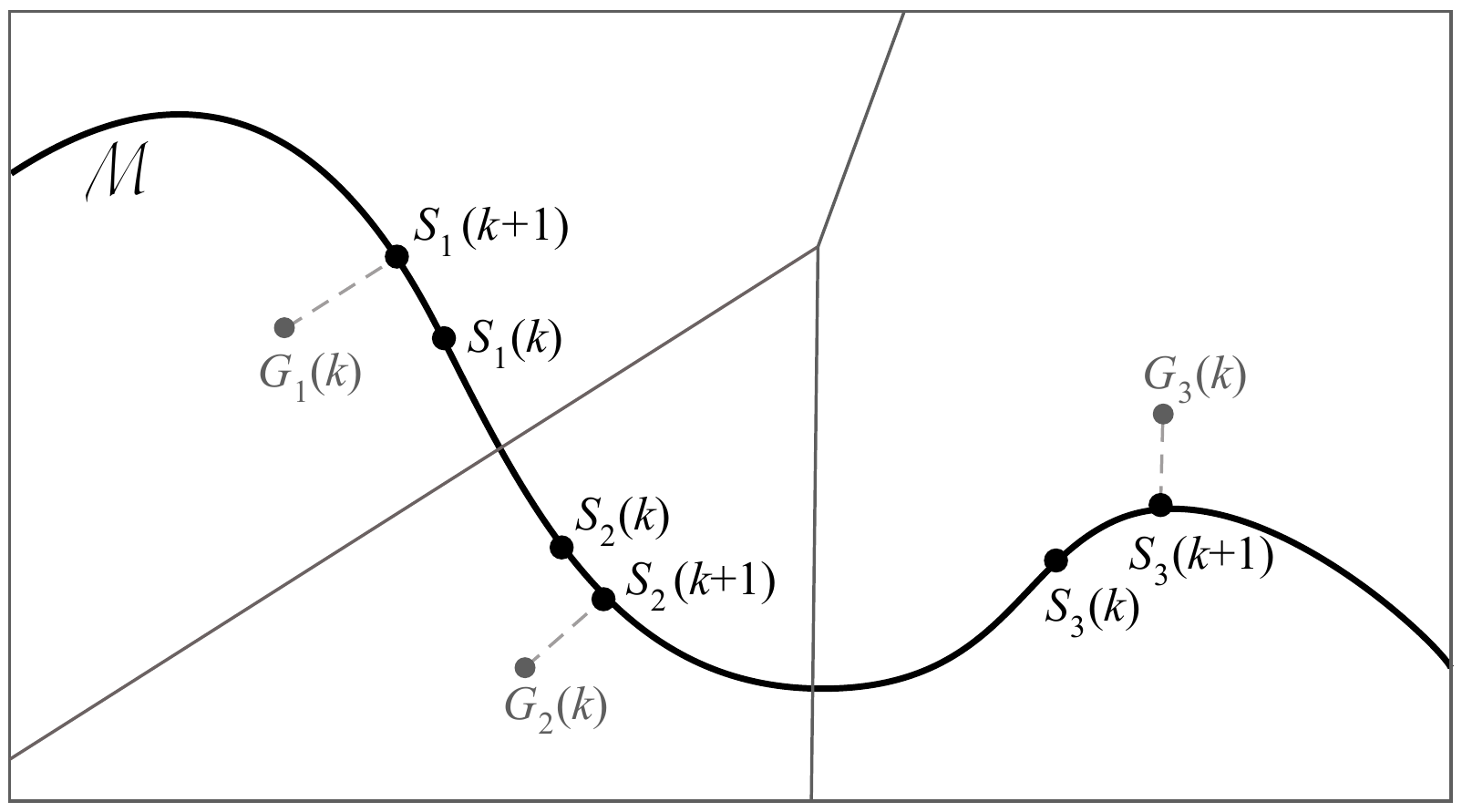}
  \caption{Illustration of a single iteration of the algorithm with three samples: Each $G_i(k)$ is the centroid of the partition region corresponding to the sample $S_i(k)$ at the $k^{th}$ iteration. The updated sample $S_i(k+1)$ is the projection of the centroid  $G_i(k)$ onto the manifold.}
  \label{fig:illustration_iteration}
\end{figure}

The following is a summary of the procedure we apply for obtaining a manifold discretization that minimizes the total manifold distance estimation error. Given the available domain of parameters and the mapping defining the manifold, we begin with an initial sample set $\mathcal{S}(0) = \{S_i(0)\}$ on the manifold, which is possibly randomly selected. We optimize the sample set iteratively. In each $k^{th}$ iteration of the algorithm, we first compute the regions $\{R_i(k)\}$ that partition $R$ with respect to the manifold samples $\{S_i(k)\}$, and then we modify each sample $S_i(k)$ individually to obtain the new sample $S_i(k+1)$  such that the manifold distance estimation error given by (\ref{eq:Ei_final_norm}) is minimized in the corresponding region. The new sample $S_i(k+1)$ is the projection of the centroid $G_i(k)$ onto the manifold. Iterations are repeated until improvements become negligible. We call this algorithm Registration-Efficient Manifold Discretization (REMD). An iteration of the algorithm is illustrated in Fig.~\ref{fig:illustration_iteration}, and the pseudocode is given in Algorithm \ref{alg:parametricManifoldDiscretization}. 

Assuming that the feasible domain $\Lambda$ of parameter vectors is compact and the mapping $U$ is bounded, for a given number of samples $N$, there exists a solution $\mathcal{S}^{*}$ that globally minimizes the total error $E$ in~(\ref{eq:E_total}). At each iteration of the discussed method, first the partition regions are updated and then the samples are readjusted, both of which are modifications that either reduce $E$ or retain it. Since the error $E$ is non-increasing throughout the iterations and is also lower bounded, the algorithm converges. However, in general the cost function is a non-convex, complicated function of the parameter vectors; therefore, the algorithm is not guaranteed to converge to the globally optimal solution. This could be mitigated by the choice of a good initial distribution of samples. For instance, in order to begin with a fair and balanced sample distribution, a preliminary stage can be added before the main iterations. Here one can impose the condition that the pairwise distance between any two samples in the ambient space or the parameter space is larger than some threshold value.

\begin{algorithm}[h]
\caption{Registration-Efficient Manifold Discretization}

\begin{algorithmic}[1]

\STATE
\textbf{Input:} \\
$\Lambda$: Feasible domain of parameter vectors\\
$U$: Mapping from parameter domain $\Lambda$ to manifold  $\mathcal{M}$\\
$N$: Number of manifold samples\\

\STATE
\textbf{Initialization:}
\STATE
Choose an initial set of manifold samples $\mathcal{S}(0) = \{ S_i(0) \} $, $i=1, \cdots, N$.
\STATE
$k=0.$

\REPEAT

\STATE
Determine the partition regions $\{R_i(k)\}$.

\STATE
Compute the centroids $\{G_i(k)\}$ of the regions.

\STATE
Update each sample $S_i(k)$ to $S_i(k+1)$, which is the projection of the centroid $G_i(k)$ on the manifold.

\STATE
$k=k+1$.

\UNTIL{The difference between $\mathcal{S}(k) $ and $\mathcal{S}(k-1)$ is insignificant}

\STATE
 $\mathcal{S}= \mathcal{S}(k)$.

\STATE
\textbf{Output}:\\
 $\mathcal{S}=\{S_i\} $: A set of manifold samples\\

\end{algorithmic}
\label{alg:parametricManifoldDiscretization}
\end{algorithm}

\section{Classification-Based Discretization of Multiple Manifolds}
\label{sec:classBasedManDisc}

We have examined above a discretization solution for a single signal manifold based on the minimization of the distance estimation error given by the approximation of the manifold by a set of samples. Now we consider the sampling problem with multiple signal manifolds. We consider that the manifold distance is computed with a discrete set of samples from each manifold, and the estimated class label of a signal is the label of the nearest manifold sample. Clearly, the accuracy of a sampling in classification is highly related to its accuracy in distance approximation. One possible solution to the multiple discretization problem is to sample each class representative manifold independently with the REMD algorithm reviewed in Section \ref{sec:distanceBasedManDisc}. Although this brings a certain improvement in the classification performance compared to baseline sampling solutions such as regular or random sampling, it fails in taking into account the geometric properties of different manifolds relative to each other. Therefore, a preferable approach to the multiple discretization problem relies in the joint discretization of all manifolds.

Furthermore, given a fixed budget for the total number of manifold samples, which can also be interpreted as a fixed computational complexity for classification, we would like to determine how many samples should be selected from each manifold such that the overall classification accuracy is maximized. The convenience of representing a manifold by a discrete sample set is highly dependent on the manifold geometry. The sample budget may thus vary for different manifolds. Moreover, in the determination of the budget allocation, the relative dependencies of the manifolds must also be taken into account. For instance, if a subgroup of manifolds are more likely to lead to misclassifications because of their internal resemblance, then it may be more preferable to allocate them a higher number of samples.

In Section \ref{ssec:classDiscrSamp}, we first analyze the classification accuracy with a multiple manifold discretization, then in Section \ref{ssec:classBasedMDAlg} we describe an iterative algorithm for sampling multiple signal manifolds that aims to improve the classification accuracy gradually. In Section \ref{ssec:classBasedMDBudget} we discuss some possible approaches for determining the allocation of the overall sample budget to different manifolds.

\subsection{Classification with Discrete Samples on Manifolds}
\label{ssec:classDiscrSamp}

The classification of signals with discretized manifolds can be formulated as follows. We consider $M$ signal manifolds $\mathcal{M}_1, \mathcal{M}_2, \cdots , \mathcal{M}_M $,  where each manifold $\mathcal{M}_m$ is approximated by the finite set

\begin{equation}
\mathcal{S}^{m} =  \{ S_i^m \}=  \{  U_m(\lambda_i^m) \}   \subset {\mathcal{M}_{m}}
\end{equation}
of $N_m$ samples, for $i=1, \cdots, N_m$ and $ \lambda_i^m  \in \Lambda_m $. The classification problem consists in determining the closest manifold for each test signal. Given a signal $x \in \mathbb R^n$, the estimation $\hat{l}(x)$ of its true class label $l(x)$ is determined by its nearest neighbour among all manifold samples:

\begin{eqnarray}
\begin{split}
\hat l(x)&= \arg \min_m D(x, \mathcal{S}^{m}) \\
&= \arg \min_m \bigg(  \min_{i=1,..,N_m} d(x,S_i^m) \bigg) .
\end{split}
\end{eqnarray}

We analyze now the classification error induced by the discretization of manifolds. Let $R \subset \mathbb R^n $ denote a bounded and piecewise connected region of interest in the signal space. For each manifold $\mathcal{M}_m$, we define a partitioning of $R$ into regions $\{R_i^m\}$, where each region consists of points closest to a specific sample $S_i^m$ of $\mathcal{M}_m$ among its all samples,
\begin{eqnarray}
\begin{split}
R&=\bigcup_{i=1}^{N_m} R_i^m, \\
R_i^m&=\{ x \in R: \, d(x,S_i^m) \leq d(x,S_j^m), \\
&\, \, \, \, \, \, \, \, \, \, \, j\in \{  1, \cdots , N_m \}  \}.
\end{split}
\end{eqnarray}
Now, consider a signal $x \in R_i^m$ that is of the class given by $\mathcal{M}_m$, i.e., $x \in R_i^m \cap \mathcal{H}_m $. Depending on the distribution of samples, $x$ can be correctly classified only if its distance to $S_i^m$ is the smallest among the distances to all other manifold samples. Thus, we define a function $E_i^m: R_i^m \rightarrow \{ 0,1 \}$ such that it represents the classification error for signals of class $m$ in the region $ R_i^m$:

\begin{equation}
E_i^m(x)=\bigg\{ 
\begin{array} {l}
1 \, \, \text{    if   } x\in \mathcal{H}_m \text{ and } \, d(x,S_i^m) > D\big(x, \bigcup_{r \neq m} \mathcal{S}^r  \big) \\ 
0 \, \, \text{    otherwise}  \end{array}
\label{eq:errorPerCell1}
\end{equation}
Then, from (\ref{eq:errorPerCell1}) we define the total classification error $E$,
\begin{equation}
E=\sum_{m=1}^M \sum_{i=1}^{N_m}  \int_{R_i^m} E_i^m(x)    \, dx.
\label{eq:errorTotalCell1}
\end{equation}
Notice that due to the definition of the error $ E_i^m(x)$, the total classification error $E$ corresponds to the sum of the volumes of the regions in $R$ where signals are not correctly classified. Another source of misclassification associated with a sample $S_i^m$ corresponds to the points that are actually closer to another manifold than $\mathcal{M}_m$, but are misclassified as a result of being closer to $S_i^m$ than their nearest manifold sample of the correct class. Hence, in analogy with $E_i^m$, we can define an alternative classification error function $F_i^m: R_i^m \rightarrow \{ 0,1 \} $ as

\begin{equation}
F_i^m(x)=\bigg\{ 
\begin{array} {l}
1 \, \, \text{    if   } x \notin \mathcal{H}_m \text{ and }  d(x,S_i^m) < D\big(x, \bigcup_{r \neq m} \mathcal{S}^r \big) \\ 
0 \, \, \text{    otherwise}  \end{array},
\label{eq:errorPerCell2}
\end{equation}
which leads to the following alternative formulation of the total classification error $F$:

\begin{equation}
F=\sum_{m=1}^M \sum_{i=1}^{N_m}  \int_{R_i^m} F_i^m(x)    \, dx.
\label{eq:errorTotalCell2}
\end{equation}

The classification errors $E$ in (\ref{eq:errorTotalCell1}) and $F$ in (\ref{eq:errorTotalCell2}) are equal. However, due to the two mentioned sources of misclassification associated with a single sample $S_i^m$, we formulate the classification error as the combination of the two, where the reason for this choice is made more clear in the algorithm description in Section \ref{ssec:classBasedMDAlg}. Hence we write the total classification error as 
\begin{equation}
\varepsilon= \frac{1}{2} \sum_{m=1}^M \sum_{i=1}^{N_m}  \int_{R_i^m} \big( E_i^m(x) + F_i^m(x) \big)   \, dx,
\label{eq:errorTotalCell}
\end{equation}
where $\varepsilon = E = F$.

\begin{figure}
 \centering
  \includegraphics[scale=0.85]{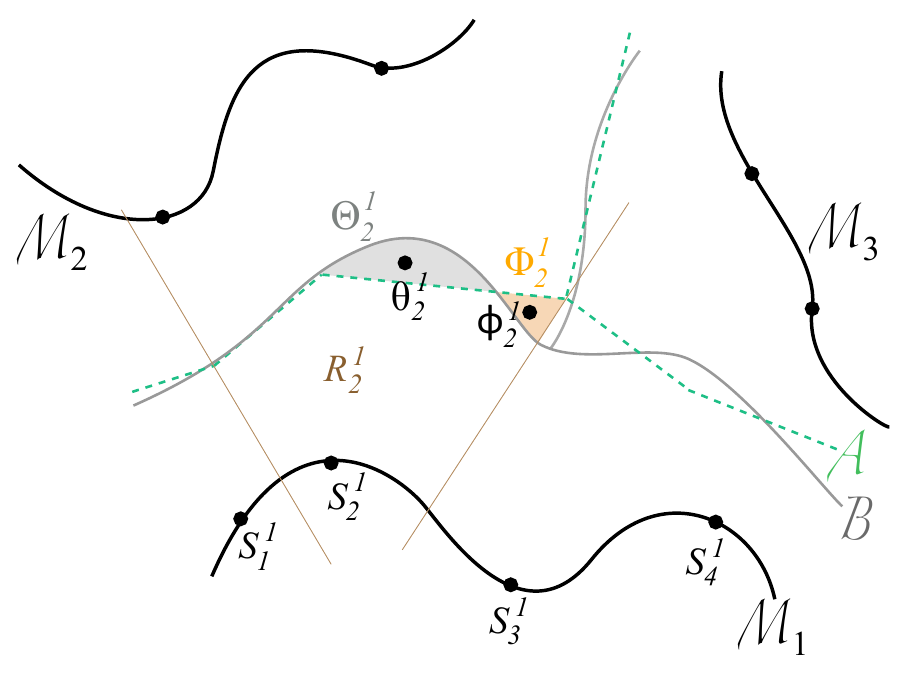}
  \caption{Illustration of transformation-invariant signal classification via transformation manifolds}
  \label{fig:illustrationManifolds2}
\end{figure}

Note that a geometric interpretation of the problem is the following. Let $\tilde R_i^m$ denote the region of space with smallest distance to the sample $S_i^m$ among the samples of all manifolds:
\begin{equation}
\begin{split}
\tilde R_i^m&=\{x \in R: d(x,S_i^m) \leq d(x,S_j^r), \\
                     & \, \, \, \, \, \,\,\, \forall r\in \{  1, \cdots ,M \}, \forall j \in \{  1, \cdots , N_r \}  \},
\end{split}
\end{equation}
where $\tilde R_i^m \subset R_i^m $. Also, let $\mathcal{A}_{ij}^{mr}$ denote the affine hyperplane which is at equal distance to the two samples $S_i^m$ and $S_j^r$ from two different manifolds $ \mathcal{M}_m$ and $ \mathcal{M}_r$. In analogy with the way that the decision surface $\mathcal{B}$ is defined in Section \ref{sec:Notations} as the boundary for the determination of the true classes $l(x)$ of signals, let now $\mathcal{A}$ denote the decision surface which determines the class label estimation $\hat{l}(x)$ of signals based on the manifold approximations given by samples. Hence, we define
\begin{equation}
\mathcal{A}=\bigcup_{m=1}^M \partial \big( \bigcup_{i=1}^{N_m}  \tilde R_i^m \big),
\end{equation}
where $\mathcal{A} \subset \bigcup_{m, r, i, j} \mathcal{A}_{ij}^{mr}$. The definition of the classification error function $\varepsilon$ in (\ref{eq:errorTotalCell}) corresponds to the total volume of the regions between the true boundary $\mathcal{B}$ and its approximation $\mathcal{A}$. Therefore, the problem of minimizing the classification error can be regarded geometrically as the selection of the sample sets $\bigcup_{m=1}^M \mathcal{S}^m$ such that the resulting $\mathcal{A}$ constitutes an accurate approximation of $\mathcal{B}$ inside $R$. This is illustrated in Fig.~\ref{fig:illustrationManifolds2}.

\subsection{Discretization Algorithm}
\label{ssec:classBasedMDAlg}

We would like to minimize the classification error $\varepsilon$ in (\ref{eq:errorTotalCell}) by optimizing the sample sets  $ \bigcup_{m=1}^M \mathcal{S}^m $. In order to achieve this, we suggest an iterative procedure as follows. We start with an initial set of samples. Then, in each iteration we optimize one manifold sample $S_i^m$ and try to reduce $\varepsilon$ by perturbing $S_i^m$. However, the dependence of the classification error on the location of a sample $S_i^m$ is fairly intricate, and it is not simple to determine the optimal sample location. Hence, in the minimization of the error, we adopt a constructive approach rather than optimal, therefore, the search directions in the perturbation of a sample may not always decrease the overall error. In order to handle this, we accept an update on a sample location only if it reduces the classification error. After reaching a locally optimum error with the perturbation of the single sample $S_i^m$, we repeat this process with different manifold samples until convergence. The reduction of the error at each step assures the convergence of the algorithm. The overall procedure is not guaranteed to converge to the globally optimal solution and results in a local minimum, whose accuracy depends on the initialization of the samples.

In a single iteration of the algorithm, we would like to find an update on $S_i^m$ that reduces the error $\varepsilon$, where the rest of the samples are considered to be fixed. The examination of Eq.~(\ref{eq:errorTotalCell}) reveals that the effect of the sample $S_i^m$ on $\varepsilon$ is twofold. The terms $E_i^m(x)$ and $F_i^m(x)$ involve the distance of space points $d(x,S_i^m)$ to the sample $S_i^m$, but the region of integration $R_i^m$ is also defined by the position of the sample $S_i^m$. Hence, the classification error has a complicated dependence on the sample location $S_i^m$. Let $\Theta_{i}^m$ and $\Phi_{i}^m$ denote the regions of $R_i^m$ where $E_i^m(x)=1$ and $F_i^m(x)=1$ respectively (see Fig.~\ref{fig:illustrationManifolds2}):

\begin{eqnarray}
\begin{split}
\Theta_{i}^m&=\{ x\in R_i^m \cap  \mathcal{H}_m: \\
                 &\, \, \, \, \, \, \, \, \, \, \, \, \, \, \,   d(x,S_i^m) > D\big(x, \bigcup_{r \neq m} \mathcal{S}^r \big) \} \label{eq:Reim} ,\\
                 \end{split}
\end{eqnarray}
\begin{eqnarray}
\begin{split}
\Phi_{i}^m&=\{x\in R_i^m \setminus  \mathcal{H}_m: \\
	      &\, \, \, \, \, \, \, \, \, \, \, \, \, \, \, \, d(x,S_i^m) < D\big(x, \bigcup_{r \neq m} \mathcal{S}^r \big)   \}\label{eq:Rfim}.
\end{split}
\end{eqnarray}  

The error terms in the expression (\ref{eq:errorTotalCell}) contributing to $\varepsilon$ are in fact the sum of the volumes of these two regions $\Theta_{i}^m$ and $\Phi_{i}^m$. Therefore, in order to reduce the error $\varepsilon$, we seek an update on $S_i^m$ that decreases the volumes of $\Theta_{i}^m$ and $\Phi_{i}^m$. Let $S_i^{m}(k)$ be the location of the sample $S_i^m$ at the $k^{th}$ iteration of the search algorithm, and let $R_i^{m}(k)$, $\Theta_{i}^{m}(k)$, and $\Phi_{i}^{m}(k)$ be defined similarly. The definitions (\ref{eq:Reim}) and (\ref{eq:Rfim}) suggest that decreasing the distance $d(x,S_i^{m}(k))$ between the sample and the points in $\Theta_{i}^{m}(k)$ reduces the misclassified portion of $\Theta_{i}^{m}(k)$. Similarly, it is necessary to increase the distance $d(x,S_i^{m}(k))$ between the sample and the points in $\Phi_{i}^{m}(k)$ in order to reduce the misclassified portion of $\Phi_{i}^{m}(k)$. Hence, we define the distance measures $D_{\Theta}(S_i^{m}(k))$ and $D_{\Phi}(S_i^{m}(k))$ as follows:

\begin{eqnarray}
D_{\Theta}(S_i^{m}(k))=\int_{\Theta_{i}^{m}(k)} d^2(x,S_i^{m}(k))  \, dx,\\
D_{\Phi}(S_i^{m}(k))=\int_{\Phi_{i}^{m}(k)} d^2(x,S_i^{m}(k))  \, dx.
\end{eqnarray}
As discussed in Section \ref{sec:distanceBasedManDisc}, the minimization of $D_{\Theta}(S_i^{m}(k))$ is possible by minimizing the distance $d(\theta_{i}^{m}(k), S_i^{m}(k) )$, where $\theta_{i}^{m}(k)$ is the centroid of $\Theta_{i}^{m}(k)$. Similarly, in order to maximize $D_{\Phi}(S_i^{m}(k))$, one should maximize $d(\phi_{i}^{m}(k), S_i^{m}(k) )$, where $\phi_{i}^{m}(k)$ is the centroid of $\Phi_{i}^{m}(k)$. However, even an update on $S_i^{m}(k) $ that decreases $D_{\Theta}(S_i^{m}(k))$ and simultaneously increases $D_{\Phi}(S_i^{m}(k))$ does not guarantee that the total classification error $\varepsilon$ decreases. This is because in general $\Theta_{i}^{m}(k+1) \not \subset \Theta_{i}^{m}(k)$ and $\Phi_{i}^{m}(k+1) \not \subset \Phi_{i}^{m}(k)$. Even if the error is reduced within $\Theta_{i}^{m}(k)$, $\Theta_{i}^{m}(k+1)$ might contain points that are not inside $\Theta_{i}^{m}(k)$ and actually increase $\varepsilon$. Still, when it is aimed to reduce $\varepsilon$ by perturbing only $S_i^{m}(k)$ in a given configuration of the samples and manifolds, curing the immediate regions of misclassification $\Theta_{i}^{m}(k)$ and $\Phi_{i}^{m}(k)$ is a promising attempt. We thus propose to update the sample $S_i^{m}(k)$ in the following way as long as the overall error does not increase. 

Let $\mu_{i}^{m}(k)$ and $\nu_{i}^{m}(k)$ denote the parameter vectors corresponding respectively to the projections of the centroids $\theta_{i}^{m}(k)$ and $\phi_{i}^{m}(k)$ on the manifold. Then, the purpose of moving $S_i^{m}(k)$ closer to $\theta_{i}^{m}(k)$ and away from $\phi_{i}^{m}(k)$ leads to the following two updates:

\begin{eqnarray}
\begin{split}
	S_i^{m}(k+1) =  U_m \big( &(1- \alpha) \lambda_{i}^{m}(k)+\alpha \, \mu_{i}^{m}(k)\big)\\
	                      & \text{ such that } \alpha \text{ minimizes } \varepsilon, \label{eq:updateSPlus}\\
\end{split}
\end{eqnarray}
\begin{eqnarray}
\begin{split}	                      
	S_i^{m}(k+1) =  U_m \big( &(1+\beta) \lambda_{i}^{m}(k) - \beta \, \nu_{i}^{m}(k) \big)\\
			  & \text{ such that } \beta \text{ minimizes } \varepsilon, \label{eq:updateSMinus}
\end{split}
\end{eqnarray}
where $\lambda_{i}^{m}(k)$ is the parameter vector defining $S_i^{m}(k)$ and both $\alpha$ and $\beta$ are positive scalars. Hence we determine the directions of perturbation with respect to the centroids of the misclassified volumes, and we adjust the amount of perturbation to obtain the largest decrease in the error. In this way, we find a locally optimum sample location reducing the misclassified portions of $\Theta_{i}^{m}(k)$ and $\Phi_{i}^{m}(k)$. It also guarantees that the possible penalty of creating new misclassified regions by moving the sample is always smaller than the benefit of correcting previous misclassifications. In the optimization of a single sample $S_i^m$, we alternate between the updates in (\ref{eq:updateSPlus}) and  (\ref{eq:updateSMinus}) until convergence, where the parameters $\mu_{i}^{m}(k)$ and $\nu_{i}^{m}(k)$  are updated after each perturbation. Then we continue the optimization process by picking other manifold samples and applying the same procedure until the classification error converges. We call this algorithm Classification-Driven Manifold Discretization (CMD) and give an overview of it in Algorithm \ref{alg:classificationBasedMD}. Finally, we note that at each iteration the perturbation of a manifold sample in the described way corresponds to a one-dimensional search in the $d$-dimensional parameter space. Although the algorithm is not guaranteed to be optimal, it offers a compromise in the performance-complexity trade-off.

\begin{algorithm}[h]
\caption{Classification-Driven Manifold Discretization}

\begin{algorithmic}[1]

\STATE
\textbf{Input:} \\
$\Lambda_m$: Feasible domains of parameter vectors\\
$U_m$: Mappings from parameter domains $\Lambda_m$ to manifolds  $\mathcal{M}_m$,  $m=1,\cdots,M$ \\
 $\bigcup_{m=1}^M \mathcal{S}^{m}(0) =   \bigcup_{m=1}^M \bigcup_{i=1}^{N_m} \{S_i^{m}(0) \}$: Initial set of manifold samples\\

\STATE
\textbf{Initialization:}
\STATE
Initialize total classification error $\varepsilon$ as defined in Eq.~(\ref{eq:errorTotalCell}).
\STATE
$k=0.$

\REPEAT

\STATE \label{state:pickSample}
Pick (possibly randomly) a manifold $\mathcal{M}_m$ and a sample $S_i^m$ from this manifold, $m \in \{1,\cdots, M\}$, $i \in \{1,\cdots, N_m\} $.

\REPEAT \label{state:computeRegions}

\STATE
Determine the misclassified region $\Theta_{i}^{m}(k) $ and the parameter vector $ \mu_{i}^{m}(k)$.

\STATE
Update $S_i^{m}(k+1) =  U_m \big( (1- \alpha) \lambda_{i}^{m}(k)+\alpha \, \mu_{i}^{m}(k)\big)$ such that  $\alpha$ minimizes $\varepsilon$.

\STATE
$k=k+1$.

\STATE
Determine the misclassified region  $\Phi_{i}^{m}(k) $ and the parameter vector $\nu_{i}^{m}(k)$.
\STATE
Update $S_i^{m}(k+1) =  U_m \big( (1+ \beta) \lambda_{i}^{m}(k) - \beta \, \nu_{i}^{m}(k)\big)$ such that  $\beta$ minimizes $\varepsilon$.

\STATE
$k=k+1$.

\UNTIL{The location of sample $S_i^{m}$ converges} \label{state:endWhile}

\UNTIL{ $\varepsilon$ converges}

\STATE
$\bigcup_{m=1}^M \mathcal{S}^m $\,= \, $\bigcup_{m=1}^M\mathcal{S}^{m}(k) $.

\STATE
\textbf{Output}:\\
$\bigcup_{m=1}^M \mathcal{S}^m $: A set of representative samples for each manifold

\end{algorithmic}
\label{alg:classificationBasedMD}
\end{algorithm}

\subsection{Sample Budget Allocation}
\label{ssec:classBasedMDBudget}

We have considered so far that the number of samples per manifold is predetermined. We address now the problem of the allocation of samples from a total budget to the different manifolds. This allocation is driven by the properties of the different manifolds. We propose two solutions for budget allocation that can be paired with the CMD algorithm.

A first simple way of determining the sample budget between manifolds is the following. We initialize the sample set of each manifold to be a dense grid on the manifold, and delete samples progressively until the number of total samples meets the budget constraint. We determine the sample to be deleted based on the classification error associated with each sample. Following the previously mentioned arguments, we delete a grid point on one manifold where $\Theta_{i}^{m}$ is relatively small. When the number of remaining grid points reaches the budget, we optimize the resulting sample sets using CMD for the final adjustment of sample locations. Since the optimization of sample locations takes places after the budget allocation, we name this approach Manifold Discretization with Predefined Allocation (MDPA).

Secondly, we introduce a joint budget allocation and CMD sample optimization solution where we allow the deletion of a sample from one manifold to create a new sample in another manifold during the iterations. In order to elaborate on the transfer of samples between manifolds, we turn back to the geometric interpretation of our problem formulation. In a configuration with multiple manifolds, our definition of the classification error is the total volume of the regions lying between the piecewisely planar boundary surface $\mathcal{A}$ formed by the sample sets and the true boundary surface $\mathcal{B}$. As a result, the classification accuracy of such a setting is directly dependent on the approximability of the boundary surface $\mathcal{B}$ by a piecewise linear model. A ``well-behaved'' manifold region corresponding to a well-approximable part of $\mathcal{B}$ is more amenable to be represented by a small set of discrete samples than a manifold region corresponding to a part of $\mathcal{B}$ that is difficult to approximate by a linear model. Therefore, the compensation of the loss of a sample from a well-behaved manifold region is relatively easier. We thus propose the following method for dynamic sample budget allocation and optimization. We start with an equal distribution of samples per manifold that satisfies the overall budget constraint, and begin optimizing them with the CMD algorithm. During iterations, we deduce that the manifold region around a sample has poor representability, whenever the misclassified region $\Theta_{i}^{m}$ has a large volume compared to average and the iterations \ref{state:computeRegions}-\ref{state:endWhile} in Algorithm \ref{alg:classificationBasedMD} fail to improve the classification accuracy. In a similar way, we determine regions of good representability around samples where the volume of the corresponding $\Theta_{i}^{m}$ is relatively small. Consequently, we add a new sample to the poorly representable region, at the cost of deleting a sample from a well representable region selected among all manifolds. We place the new sample at the point corresponding to the projection of the centroid of $\Theta_{i}^{m}$ onto the manifold. We further adapt the discretization to the new configuration in the regions where sample deletion and creation have taken place, and we apply the iterations \ref{state:computeRegions}-\ref{state:endWhile} of CMD to all neighboring samples on the same manifolds as the deleted sample and the newly created sample. Note however that the sample transfer is not guaranteed to reduce the classification error. In order to ensure the improvement of the classification accuracy, we finally accept the update only if it reduces the error. We call this approach Dynamic Manifold Discretization (DMD).

\section{Experimental Results}
\label{sec:ExpRes}

\subsection{Setup}
\label{ssec:ExpSetup}

We now present experimental results demonstrating the performances of the manifold discretization algorithms discussed in Sections \ref{sec:distanceBasedManDisc} and \ref{sec:classBasedManDisc}. All experiments are conducted on two different kinds of image appearance manifolds, namely the transformation manifold of a 2D pattern, and the manifold generated by the observations of a synthetical 3D object under varying viewpoint.

\begin{figure}[t]
 \centering
  \includegraphics[width=9cm]{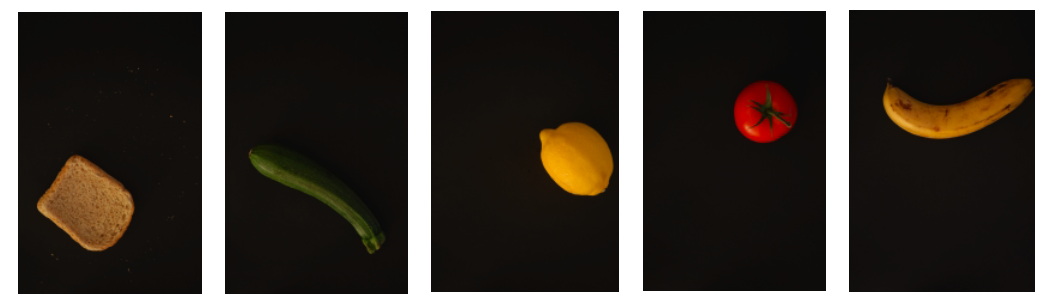}
  \caption{Example images from database}
  \label{fig:database_food}
\end{figure}

\begin{figure}
\begin{center}
     \subfigure[]
       {\includegraphics[width=1.5cm]{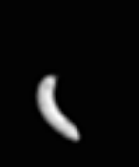}}
     \subfigure[]
       {\includegraphics[width=1.5cm]{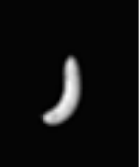}}
     \subfigure[]
       {\includegraphics[width=1.5cm]{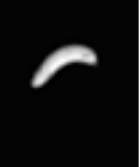}}    
     \subfigure[]
       {\includegraphics[width=1.5cm]{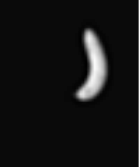}}                       
 \end{center}
 \caption{Images from a pattern transformation manifold}
 \label{fig:manifoldBanana}
\end{figure}

In the first experimentation setup, we construct pattern transformation manifolds generated by the 2D rotation and translation of visual patterns. Given a pattern $p$, we define its transformation manifold $\mathcal{M}$ by
\begin{equation} \mathcal{M} = \{U(\lambda)p: \lambda=(\psi, t_x, t_y) \in \Lambda  \}, \label{eq:patternTrModel} \end{equation}
where $\psi $ is the rotation parameter, $ t_x$ and $ t_y $ are the horizontal and vertical translation parameters, and $\Lambda$ is the domain of transformation parameter vectors. For experimentation, we use a database of top-view images of 5 different objects, where each object has 500 different images captured under different orientations and positions. An example image for each object is displayed in Fig.~\ref{fig:database_food}. Note that due to the positioning of the camera and the limitations on object positions, the 2D pattern transformation model in (\ref{eq:patternTrModel}) constitutes an approximate model for the observations. For each of the objects we build the transformation manifold of a fixed representative pattern picked among the database images. The image set of each object is grouped randomly into 300 training and 200 test images. The categorization of the database into training and test sets is changed randomly at each repetition of the experiment. For computational convenience, all images are converted to greyscale, downsampled to a resolution of 50$\times60$ pixels, and background pixels are set to the luminance value of 0 by simple thresholding. Manifold points are generated by rotating and translating the representative pattern (cropped previously near the boundary) over a 50$\times60$ pixel zero background within the parameter range $\psi \in [-\pi,\pi)$; $t_x \in [-7, 7]$; $t_y \in [-12, 12]$. All images and generated manifold points are normalized to have unit norm. Some illustrative images from the transformation manifold of one of the objects are given in Fig.~\ref{fig:manifoldBanana}. 

\begin{figure}
 \centering
  \includegraphics[scale=0.14]{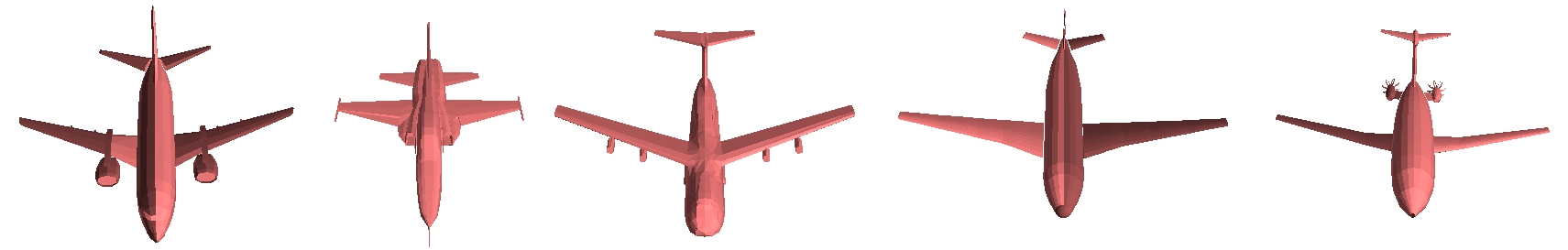}
  \caption{Example objects from the airplane class}
  \label{fig:database_proj3d}
\end{figure}

\begin{figure}
\begin{center}
     \subfigure[]
       {\includegraphics[width=1.5cm]{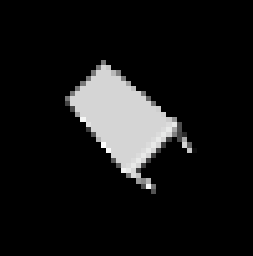}}
     \subfigure[]
       {\includegraphics[width=1.5cm]{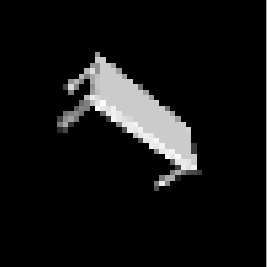}}
     \subfigure[]
       {\includegraphics[width=1.5cm]{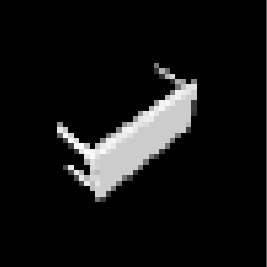}}    
     \subfigure[]
       {\includegraphics[width=1.5cm]{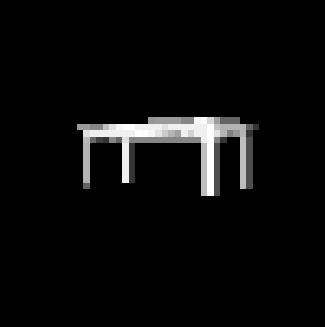}}                       
 \end{center}
 \caption{Images from an object observation manifold}
 \label{fig:manifoldTable}
\end{figure}

In the second experimentation setup, the object observation manifolds are generated as follows: Given a synthetical 3D object model $P$, we consider the observation manifold defined by
\begin{equation} \mathcal{M} = \{U(\lambda)P: \lambda=(\psi_x,\psi_y,\psi_z) \in \Lambda  \}, \label{eq:objObsModel} \end{equation}
where $U(\lambda)P$ is the image of the object rendered under the viewpoint specified by the three rotation angles $\psi_x,\psi_y,\psi_z$. We use the Princeton Shape Benchmark database of 3D models \cite{PrincetonShape}, where we conduct our experiments on 8 different classes of objects (car, airplane, ship, tank, human, animal, table, bottle) with several (4-30) objects belonging to each class. Some example objects belonging to the airplane class are shown in Fig~\ref{fig:database_proj3d}. For each class we choose a representative object, and generate the observation manifold of the representative object in the parameter range $\psi_x, \psi_y,\psi_z \in [-\pi/4, \pi/4]$. The representative object of each class is changed randomly in repetitions of the experiment. All rendered images are converted to greyscale, downsampled to the resolution of 50$\times$50 pixels and normalized to unity. The training and test sets for each manifold consist of 500 random observations of the objects of the same class within the same parameter range. Some images from an object observation manifold are displayed in Fig.~\ref{fig:manifoldTable}.

In both experimental setups, we use training images only for the computation of the centroids of space regions. We compute the centroid of a region of $ \mathbb R^n$ experimentally by taking random training images, checking if they are in the inquired region, and then computing the arithmetic average of inliers when a sufficient number of them are accumulated as suggested in \cite{128857}. After the centroids are computed, we estimate their projections onto the manifold with the aid of a dense grid on the manifold. We first locate the projection coarsely by finding the grid point that has the smallest distance to the centroid, and then refine the location of the projection by minimizing its distance to the centroid using gradient descent tools.

\begin{figure}
\begin{center}
     \subfigure[Registration errors given by the samplings]
       {\label{fig:reg_error_food}\includegraphics[width=7.5cm]{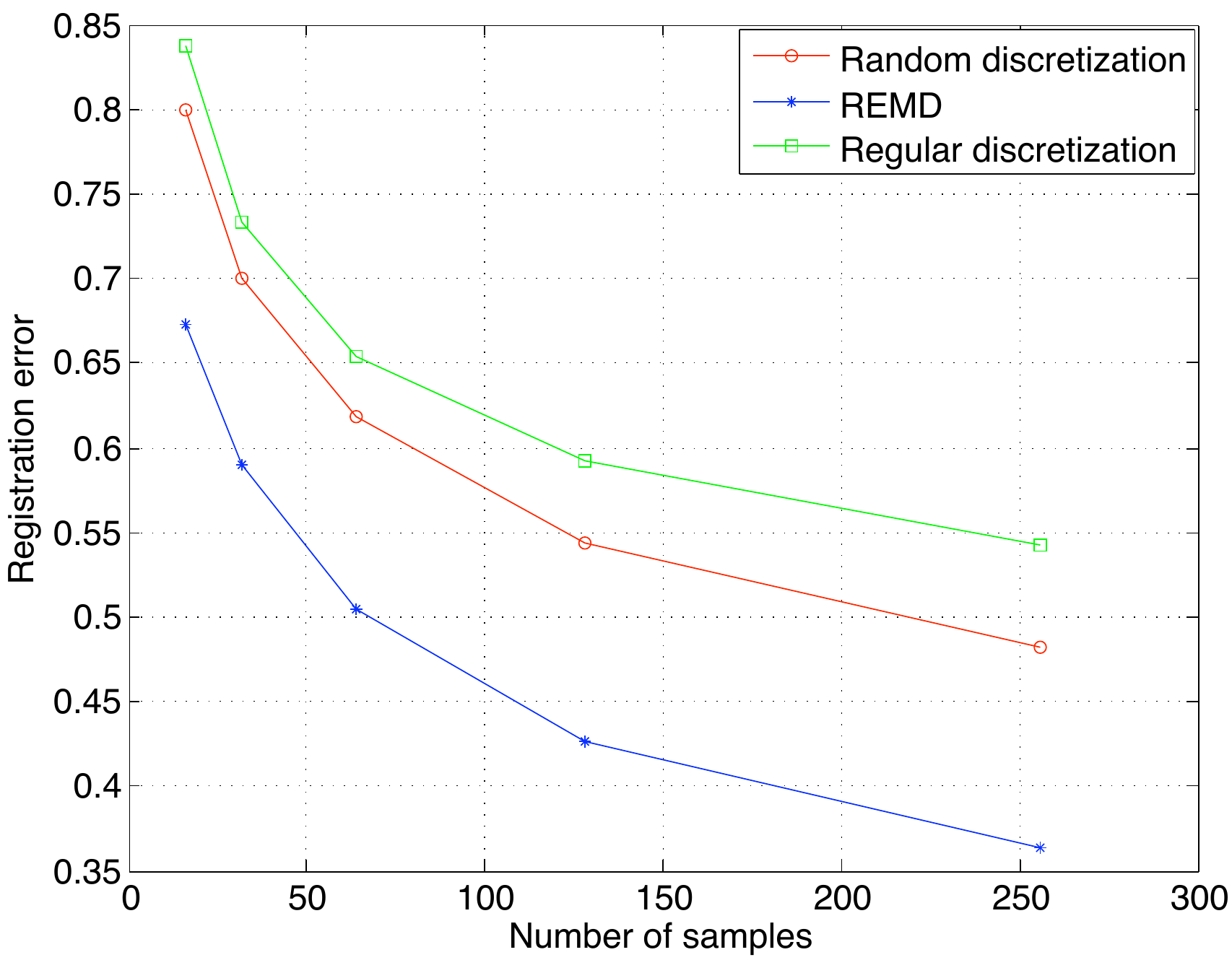}}
      \subfigure[Distribution of test points with respect to projection accuracy]
       {\label{fig:count_food}\includegraphics[width=7.5cm]{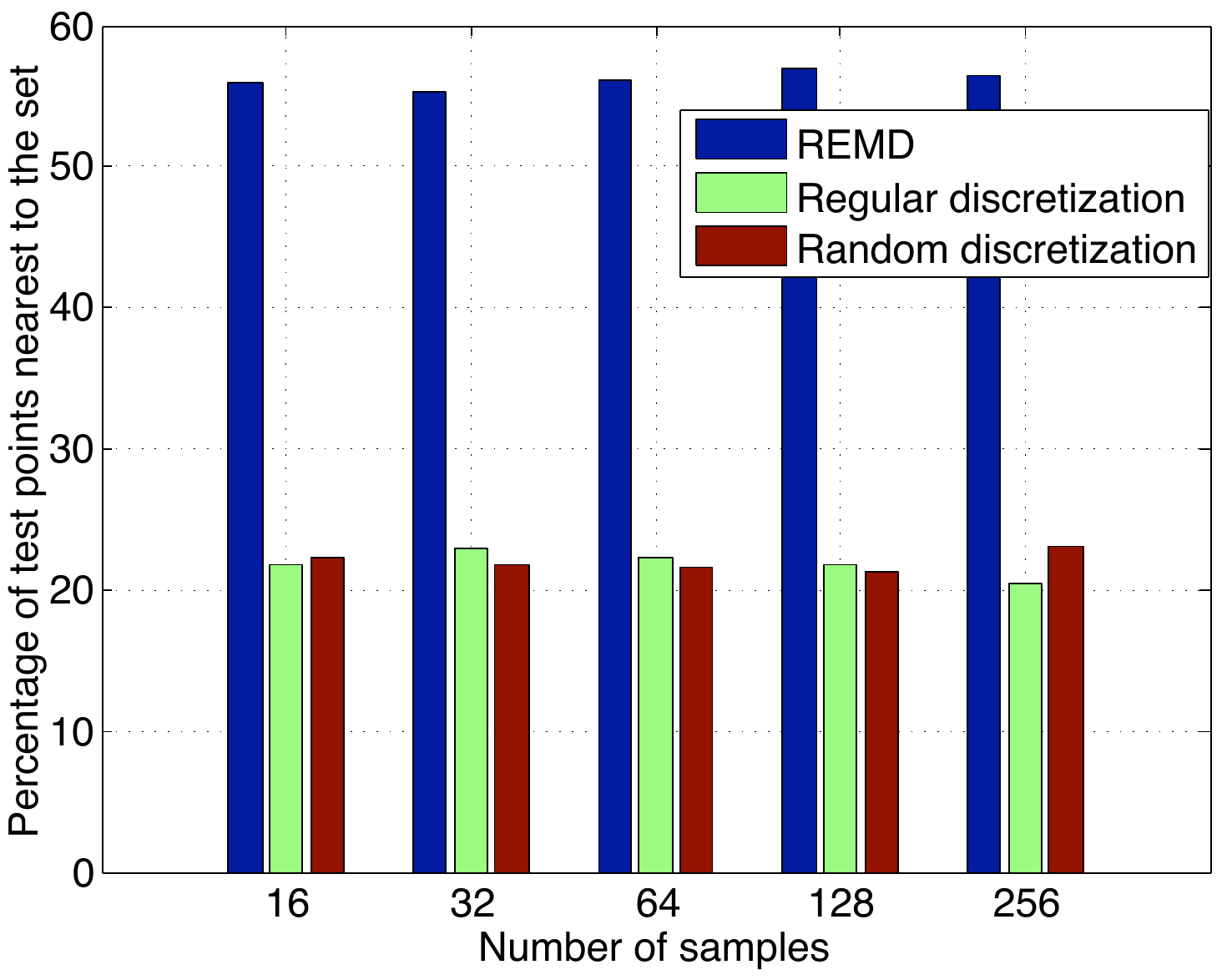}}             
 \end{center}
 \caption{Sampling results obtained on pattern transformation manifolds}
 \label{fig:results_regalg_food}
\end{figure}

\begin{figure}[t]
\begin{center}
     \subfigure[Registration errors given by the samplings]
       {\label{fig:reg_error_proj3d}\includegraphics[width=7.5cm]{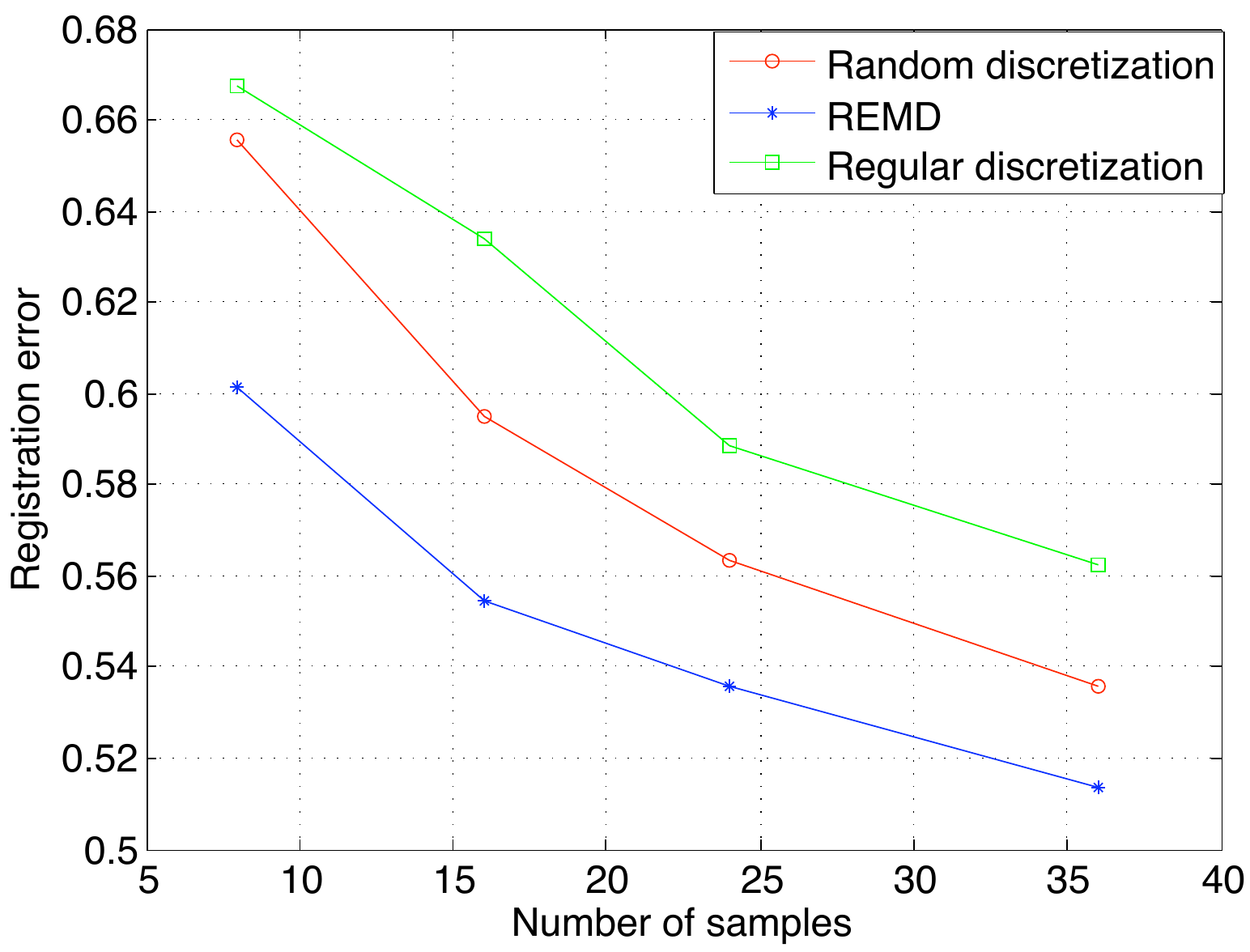}}
      \subfigure[Distribution of test points with respect to projection accuracy]
       {\label{fig:count_proj3d}\includegraphics[width=7.5cm]{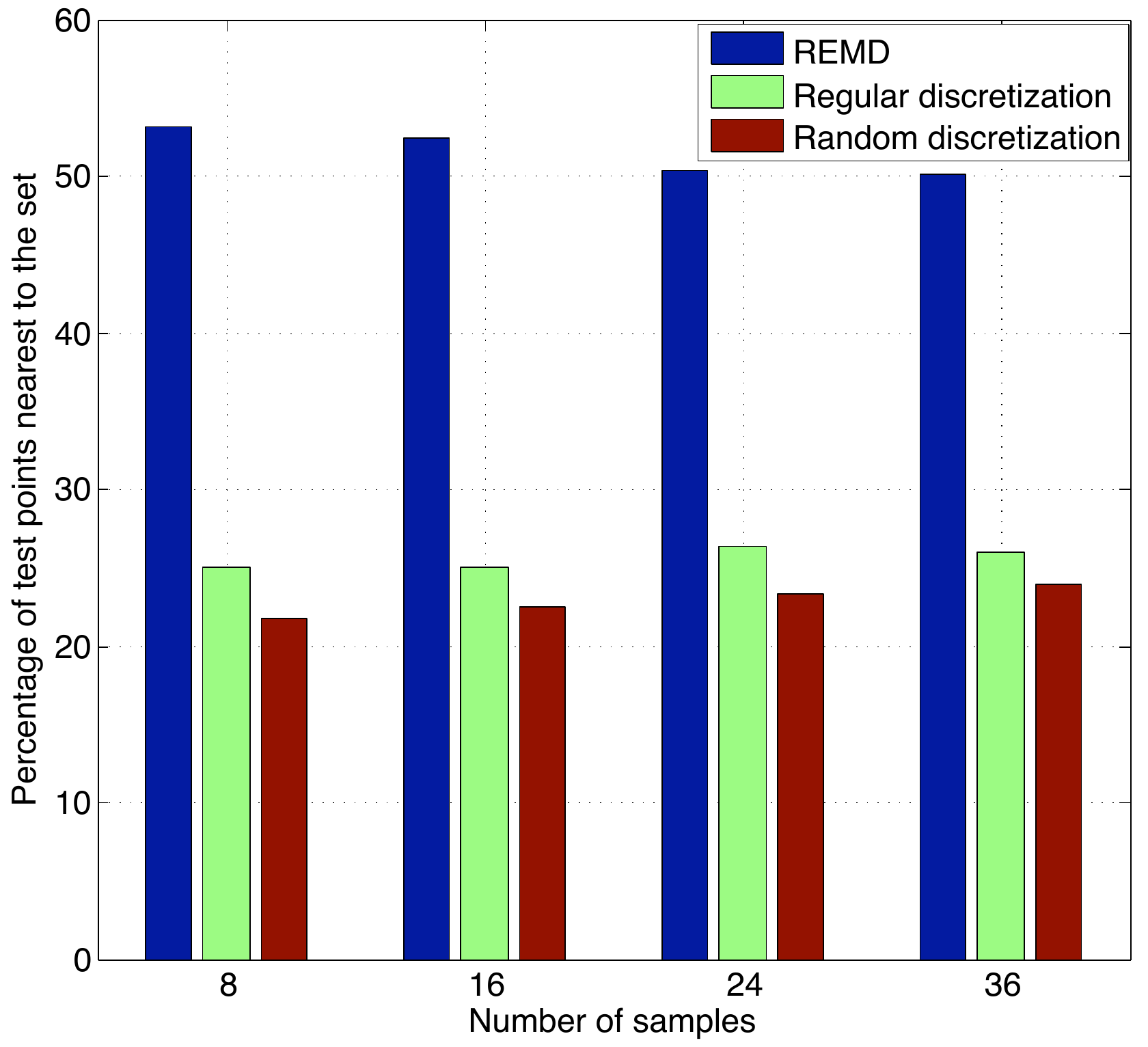}} 
 \end{center}
 \caption{Sampling results obtained on object observation manifolds}
 \label{fig:results_regalg_proj3d}
\end{figure}

\subsection{Results on registration accuracy}
\label{ssec:ExpResRegistration}

Here we test the REMD algorithm on pattern transformation and object observation manifolds. In both experimental setups, we initialize the algorithm with a randomly selected sample set. We compare the sample set determined by the REMD algorithm to the initial random sample set and to the sample set given by a regular grid over the parameter domain. The performance evaluation criterion is the accuracy of the discretization in manifold distance estimation. For each discretization, the distances of test points to the sample set are computed and the average registration error is calculated. The registration error is taken as the $\ell_2$-distance between the exact projection of the test point onto the manifold and the manifold sample with smallest $\ell_2$-distance to the test point.

In the setup with pattern transformation manifolds, we build and sample the transformation manifold of each object individually. For each object the experiment is repeated 10 times with different random initializations. The results are averaged over all realizations and all objects. In Fig.~\ref{fig:reg_error_food}, average registration errors obtained with the REMD output, random and regular sample sets are plotted for several numbers of samples. Then, we compute the distance of each test point to all manifold samples obtained with these three discretization approaches. For each test point we determine the sample among all three sets that has the smallest distance to the test point. In Fig.~\ref{fig:count_food} we report the percentage of the test points that have their closest manifold sample within the REMD output, random and regular sample sets, respectively.
 
These experiments intend to measure the capability of the discretization to provide an accurate approximation of the projection onto the manifold. As shown in the figures, the discretization obtained by the REMD method yields the least registration error when compared to the random discretization and the regular discretization in the parameter domain. In addition, for the majority of the test points the most accurate approximation of the projection lies within the REMD algorithm output sample set.

The experiments on object observation manifolds are conducted similarly. We construct and sample the observation manifold of each object individually. Experiments are repeated 5 times for each class with different random initializations. The results are again averaged over all realizations and all objects. Fig.~\ref{fig:reg_error_proj3d} shows the registration errors obtained with the samplings and Fig.~\ref{fig:count_proj3d} shows the percentage of test images with the best projection approximations within the REMD output, random and the regular sample sets. The results are in accordance with the results obtained on pattern transformation manifolds. Note that although the intrinsic dimensions of manifolds are the same in the two experimental setups, the typical number of samples required for accurately representing the pattern transformation manifolds and the object observation manifolds is quite different in these experiments. This is due to the differences in the type and range of geometric transformations that generate the manifolds. The fact that the object observation manifolds are defined over a relatively small parameter domain makes it possible to represent them with fewer samples compared to the pattern transformation manifolds for similar performance.

\subsection{Results on transformation-invariant classification}
\label{ssec:ExpResClassification}

In this part, we evaluate the performances of the discretization approaches in Section \ref{sec:classBasedManDisc} in transformation-invariant classification. In all of the pattern transformation manifold experiments, the image set of each object in the database is regarded as a different signal class. Similarly, in the synthetical objects database, the rendered images of each class of objects are considered to belong to a separate signal class. Only training images are available to the sampling algorithms, and the classification performance is measured on test images. Once the sample sets are obtained, the classification of a test image is performed by assigning the label of the manifold sample with smallest distance. In all of the following figures, the correct classification rates of test images are plotted in percentage with respect to the number of samples per manifold. All experiments are repeated 10 times with different random algorithm initializations and averaged.

\begin{figure}
\begin{center}
     \subfigure[Pattern transformation manifolds]
       {\label{fig:classIndivFood}\includegraphics[width=7.5cm]{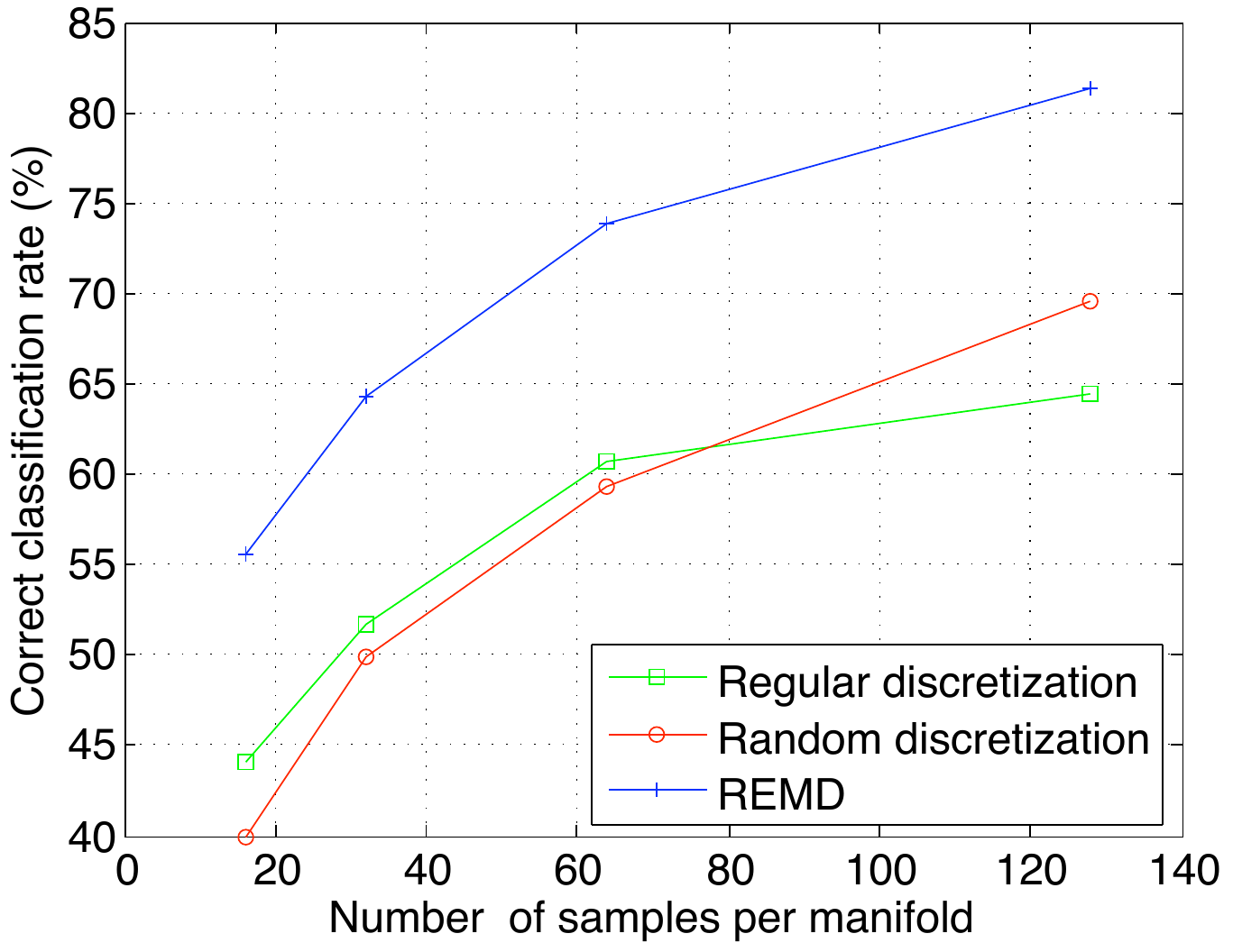}}
      \subfigure[Object observation manifolds]
       {\label{fig:classIndivProj3d}\includegraphics[width=7.5cm]{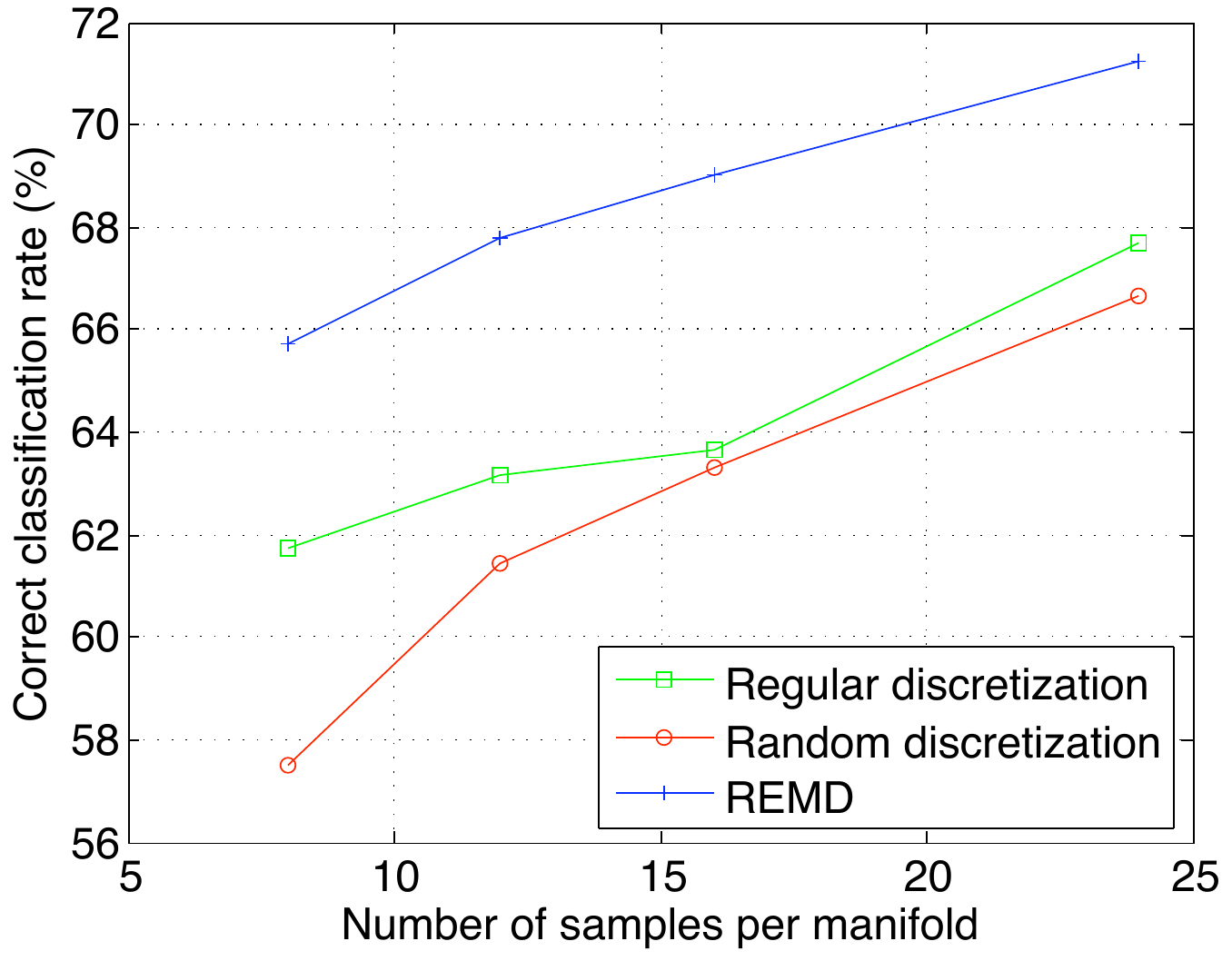}} 
 \end{center}
 \caption{Classification results obtained by sampling class representative manifolds individually}
 \label{fig:results_indiv}
\end{figure}

First, we compare the REMD algorithm, the random discretization and the regular discretization in the parameter domain with respect to their classification performances. The experimental setting is the same as that of Section \ref{ssec:ExpResRegistration}, i.e., the transformation manifold of each class is sampled individually via the REMD algorithm, randomly and regularly, where an equal number of samples are selected on each manifold.  The results obtained on pattern transformation manifolds and object observation manifolds are plotted respectively in Fig.~\ref{fig:classIndivFood} and Fig.~\ref{fig:classIndivProj3d}. The plots indicate that in both setups, the REMD output sample set has higher classification performance compared to the random and regular discretizations. This is in agreement with the results of the registration experiments of Section \ref{ssec:ExpResRegistration}, confirming the dependency of the transformation-invariant classification performance on the accuracy of manifold distance estimation. When the two plots in Fig.~\ref{fig:classIndivFood} and Fig.~\ref{fig:classIndivProj3d} are compared, it is seen that the classification rate improvement introduced by REMD or by increasing the number of samples is higher in pattern transformation manifolds. This can be explained by the difference between the two setups. In object observation manifold experiments there are several object models belonging to the same class. Therefore, space points have a relatively large deviation from the manifold of the representative object. This deviation is smaller in pattern transformation manifolds as the space points of a specific class are the images of the same object.

\begin{figure}
\begin{center}
     \subfigure[Pattern transformation manifolds]
       {\label{fig:classJointFood}\includegraphics[width=7.5cm]{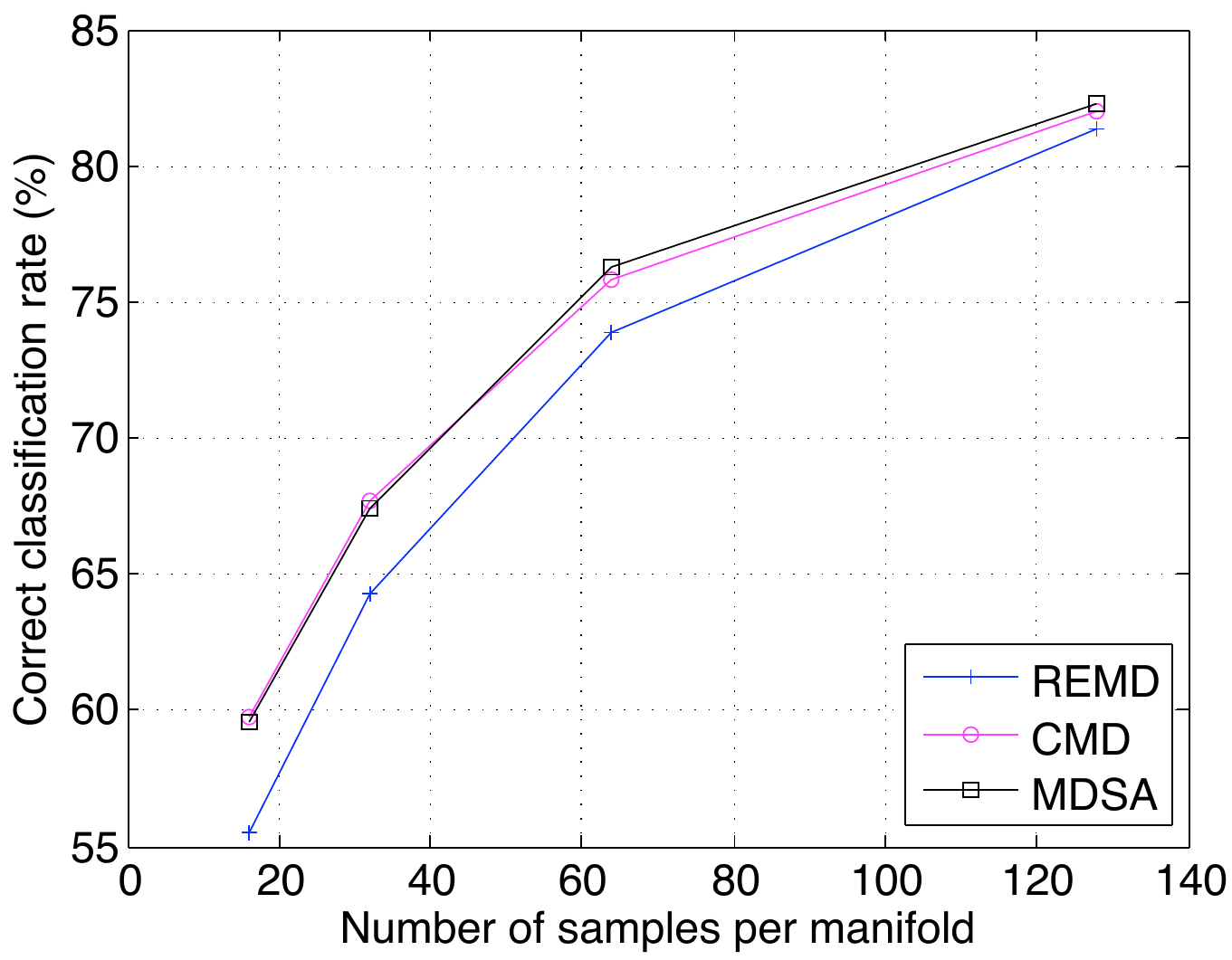}}
      \subfigure[Object observation manifolds]
       {\label{fig:classJointProj3d}\includegraphics[width=7.5cm]{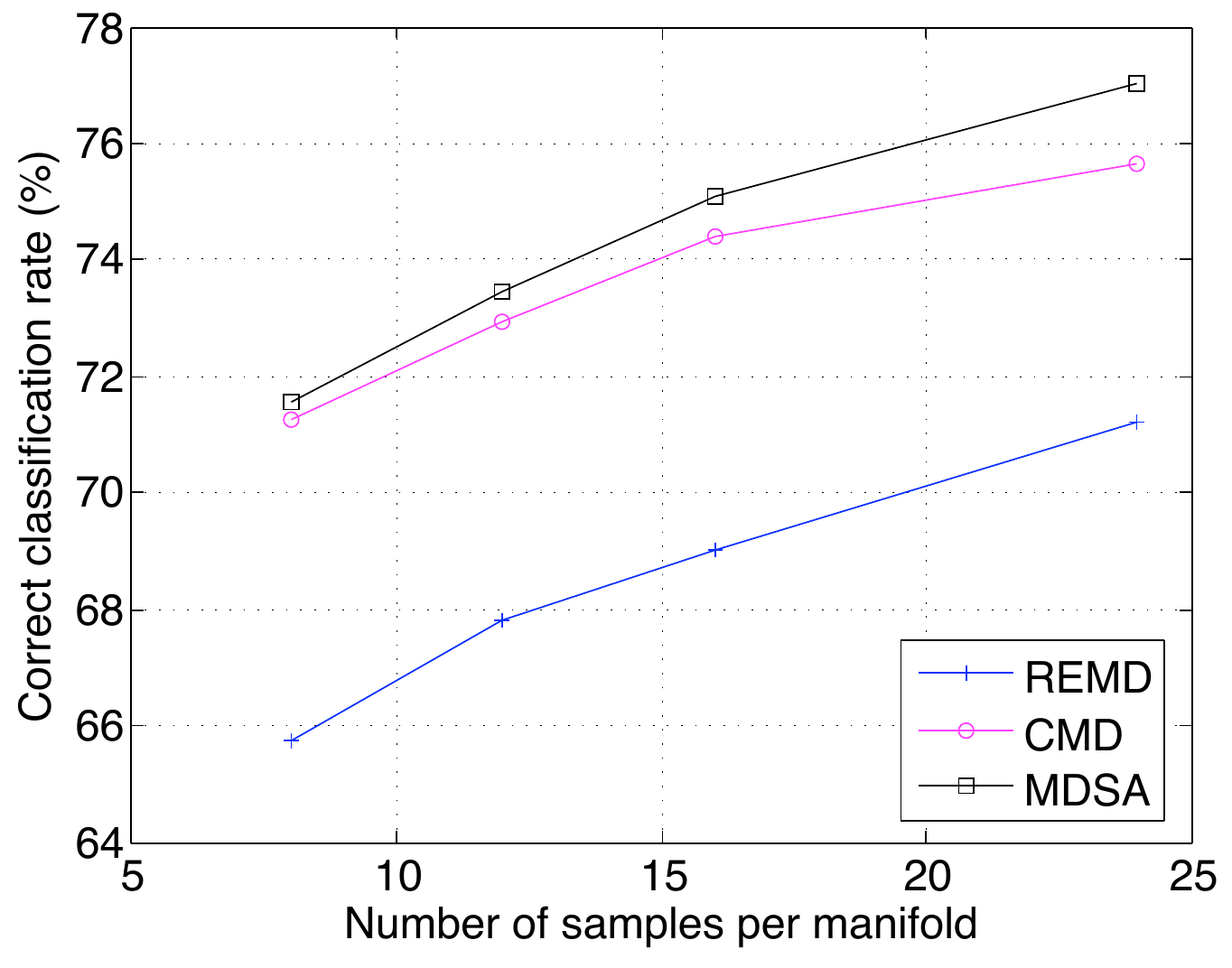}} 
 \end{center}
 \caption{Effect of the joint optimization of samples on classification accuracy}
 \label{fig:results_joint}
\end{figure}

Then we search the efficiency of jointly optimizing all manifold samples in comparison with sampling each manifold individually. For this purpose, we first select an equal number of samples from each manifold independently with the REMD algorithm as in the previous experiment. Then we apply an additional stage of joint optimization, where we optimize the samples from all classes together via two alternative approaches. In the first approach, we optimize the output sample set of the REMD algorithm further with the CMD algorithm (marked simply as CMD in the plots). In the second one we again begin with the REMD output sample set, but then perform the joint optimization of manifold samples using a simulated annealing algorithm where we define the cost function as the classification error $\varepsilon$ in (\ref{eq:errorTotalCell}). We name this approach Manifold Discretization with Simulated Annealing (MDSA). The simulated annealing algorithm is based on seeking the global optimum by trying random search directions, whereas the CMD algorithm has a more restricted search domain. Therefore, the results with MDSA are provided as a benchmark for the evaluation of the efficiency of CMD. The classification rates are plotted in Fig.~\ref{fig:classJointFood} for pattern transformation manifolds and in Fig.~\ref{fig:classJointProj3d} for object observation manifolds. The results show that the joint optimization of manifold samples after the individual sampling stage brings a significant improvement on the classification rate. This is consistent with the expectation that the relative characteristics of manifolds, i.e., their structures with respect to each other, should also be taken into account as well as their individual characteristics in classification. Moreover, the performances of the sample sets obtained by CMD and MDSA are close to each other. This shows that CMD is an effective constructive algorithm. The slight superiority of MDSA to CMD is justifiable in the sense that the CMD algorithm performs a one-dimensional search in the parameter domain at each update step, whereas in simulated annealing the search space is full dimensional in the parameter domain.

\begin{figure}
\begin{center}
     \subfigure[Pattern transformation manifolds]
       {\label{fig:classBudgetFood}\includegraphics[width=7.5cm]{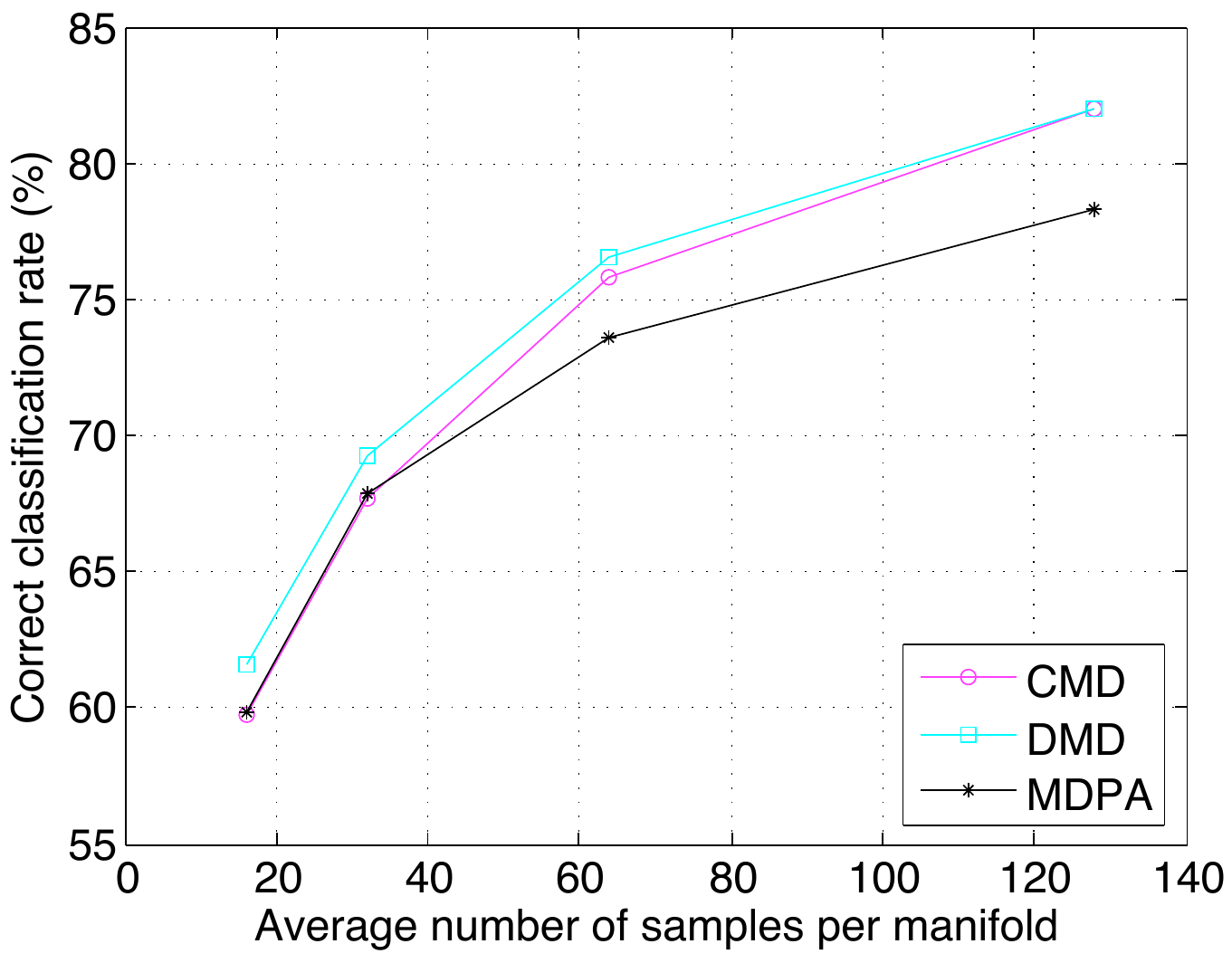}}
      \subfigure[Object observation manifolds]
       {\label{fig:classBudgetProj3d}\includegraphics[width=7.5cm]{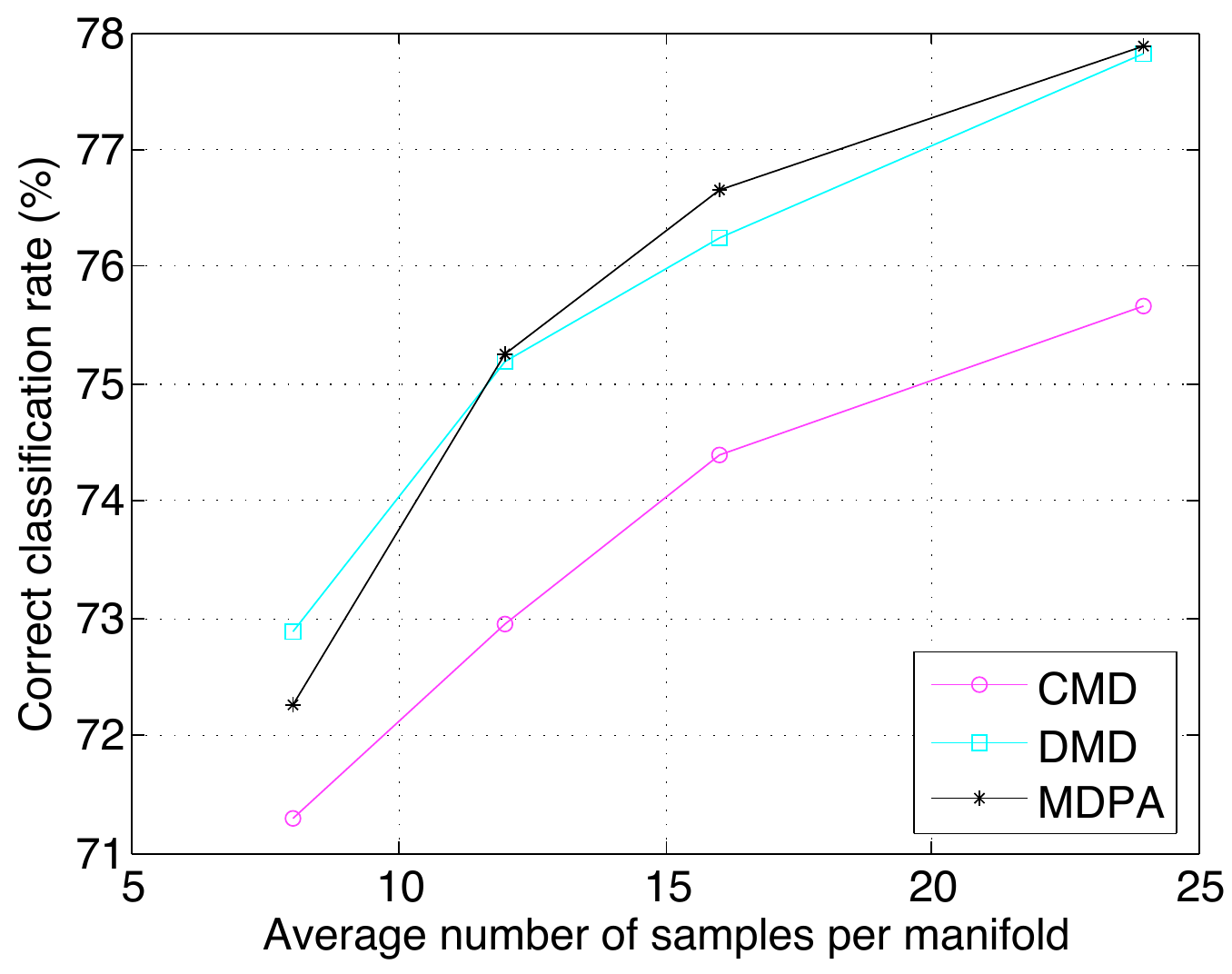}} 
 \end{center}
 \caption{Effect of the uneven distribution of sample budget on classification accuracy}
 \label{fig:results_budget}
\end{figure}

Finally, in a third experiment we examine the effect of the uneven distribution of the total sample budget to different manifolds. We compare the performances of the CMD algorithm with equal budget distribution to different manifolds, DMD and MDPA, which have been discussed in Section \ref{ssec:classBasedMDBudget}. In order to test CMD, we first select an equal number of samples from each manifold independently with the REMD algorithm, and then optimize the output of REMD in a further stage with CMD as in the previous experiment. We apply the same procedure for DMD as well; it is initialized with the output of REMD with equally distributed samples, where the movement of samples between different manifolds is allowed afterwards to optimize the distribution of the sample budget. The correct classification rates obtained with the three sampling approaches are plotted with respect to the average number of samples selected per manifold in Fig.~\ref{fig:classBudgetFood} and Fig.~\ref{fig:classBudgetProj3d}, respectively for pattern transformation and object observation manifolds. The results suggest that an uneven distribution of the sample budget to different manifolds according to their geometric properties may improve the classification accuracy when compared to the equal distribution of samples. It is seen that the performances of DMD and MDPA are close to each other in the object observation manifolds experiment. However, in the pattern transformation manifolds experiment, the number of available training images per manifold sample is much smaller, which has a negative influence on the efficiency of budget distribution in the first stage of progressive sample deletion in MDPA.

\subsection{Discussion of Results}
\label{ssec:expComputCompx}
Here we make an interpretation of our experimental results from the perspective of the trade-off between computational complexity and performance. The main motivation behind this work is the difficulty of the exact computation of the manifold distance. The state-of-the-art methods accomplishing manifold distance computation are considerably demanding. For instance, the algorithm proposed in \cite{10.1109/TPAMI.2008.156} involves a complexity of $O(K\cdot n_1 \cdot n_2)$, where $K$ is the number of atoms used in the decomposition of the reference pattern and $n_1\times n_2$ is the image resolution, while a similar algorithm complexity is reported in \cite{DBLP:journals/tmm/VasconcelosL05}. On the other hand, in order to estimate the manifold distance we propose the utilization of a suitable manifold grid that is to be determined offline. Once the grid is obtained, the manifold distance estimation is simplified merely to the computation of the norms of the difference vectors between the query signal and the samples, which clearly reduces the cost of distance estimation significantly. Meanwhile, it is not easy to draw a general conclusion in the comparison of the accuracies of manifold distance computation algorithms and our grid approach. This is highly dependent on the algorithms under comparison. For instance, the algorithm in \cite{10.1109/TPAMI.2008.156} is guaranteed to find the global solution in the projection of query images onto pattern transformation manifolds, resulting in a perfectly accurate distance computation. However, methods such as \cite{Fitzgibbon03}, \cite{DBLP:journals/tmm/VasconcelosL05}, \cite{668381} do not have such optimal performance guarantees. In the approximation of the manifold distance with a grid, the registration accuracy is clearly dependent on the number of samples, and the registration error asymptotically approaches zero as the number of samples increases. We also note that, in presence of large geometric transformations the grid approach may have an advantage over algorithms based on tangent distance, which are susceptible to local minima. Considering these, we conclude that sample-based approaches achieve a compromise between accuracy and computational effort.

Now, let us turn to the complexity of the discretization process. To start with, the REMD algorithm is in principle an adaptation of the LBG vector quantization algorithm \cite{1094577} to the manifold-modeled signal case. Therefore the main complexity of the algorithm is $O(N)$, where $N$ is the number of selected samples. As CMD is an extension of REMD to multiple manifolds, it has the same order of complexity with respect to the total number of samples. In the case that the centroids are estimated via training images, the complexity of the discretization algorithm also has a first-order dependence on the number of training images \cite{bb2675}. Also, a practical matter of concern is the estimation of the projections of the centroids onto manifolds, which we overcome by the aid of a dense grid. However, the manner of the determination of projections is out of the scope of this work, as a variety of solutions may be applicable depending on the type of the treated signals.

Finally, we note that the classification performance of CMD, which has a refined search space, is fairly close to that of MDSA where the correct classification rate is optimized by simulated annealing. Comparing these two approaches with respect to their convergence rates, we have seen that CMD terminates in a much less number of iterations, which is a result of its capability of assessing the proper search directions. Yet, the overall running time of the CMD algorithm depends on the computational time required by the projection of space points onto manifolds. The speed of the required registration block depends on the type of transformations involved. Efficient solutions exist for certain geometric transformations. For instance, the phase correlation method \cite{bb49186} is a well-known and fast technique that recovers image translations. We remark also that image registration is an active research field and recent works such as \cite{Matungka2009} are promising for the generalization of such techniques to handle a wider range of geometric transformations.

\section{Conclusion}
\label{sec:Conclusion}
We have studied the sampling of signal manifolds with known parameterization. We present a discretization solution for a single manifold based on registration accuracy. Then  we generalize the problem to the discretization of multiple signal manifolds that represent different classes of signals, and propose a method for the joint optimization of all manifold samples for the improvement of classification performance. We also discuss possible ways of optimizing the distribution of a fixed sample budget to different class representative manifolds in order to improve the classification accuracy. We test the proposed sampling approaches on pattern transformation manifolds and object observation manifolds. Experimental results indicate that the registration accuracy of the distance-based sampling is considerably higher than random and regular samplings. The consideration of the relative structures of different manifolds in the discretization improves the classification performance significantly when compared to the independent discretization of each manifold. Moreover, distributing the total sample budget unequally to the manifolds in a manner that takes account of their different characteristics may also bring an improvement. The results reveal the potential of our work to find various application areas in the treatment of parametrizable signal sets.

\section{Acknowledgement}
\label{sec:Acknowledge}
The authors would like to thank Ozan \c{S}ener for his contribution in the preparation of the experimental setup with synthetical 3D objects.

\ifCLASSOPTIONcaptionsoff
  \newpage
\fi

\bibliographystyle{IEEEtran}
\bibliography{refs}

\begin{IEEEbiography}[{\includegraphics[width=1in,height=1.25in,clip,keepaspectratio]{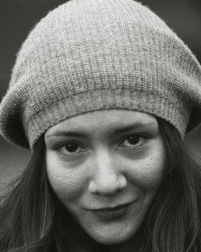}}]{Elif Vural} received the B.Sc. degrees in electrical and electronics engineering and in mathematics from Middle East Technical University (METU), Ankara, Turkey, in 2006 and the M.Sc. degree from METU in 2008 with an M.Sc. thesis on sparse 3-D scene reconstruction under the supervision of Prof. Ayd\i n Alatan. She is currently continuing the Ph.D. degree in the Signal Processing Laboratory - LTS4 at EPFL under the supervision of Prof. Pascal Frossard. She joined the Swiss Federal Institute of Technology (EPFL), Lausanne, Switzerland, as a doctoral student in 2008. Her research interests include image analysis, pattern classification, low-dimensional data representations, and multi-view geometry.
\end{IEEEbiography}

\begin{IEEEbiography}[{\includegraphics[width=1in,height=1.25in,clip,keepaspectratio]{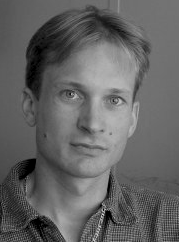}}]{Pascal Frossard} received the M.S. and Ph.D. degrees, both in electrical engineering, from the Swiss Federal Institute of Technology (EPFL), Lausanne, Switzerland, in 1997 and 2000, respectively. Between 2001 and 2003, he was a member of the research staff at the IBM T. J. Watson Research Center, Yorktown Heights, NY, where he worked on media coding and streaming technologies. Since 2003, he has been a professor at EPFL, where he heads the Signal Processing Laboratory (LTS4). His research interests include image representation and coding, visual information analysis, distributed image processing and communications, and media streaming systems. 

Dr. Frossard has been the General Chair of IEEE ICME 2002 and Packet Video 2007. He has been the Technical Program Chair of EUSIPCO 2008, and a member of the organizing or technical program committees of numerous conferences. He has been an Associate Editor of the IEEE TRANSACTIONS ON MULTIMEDIA (2004-2010), the IEEE TRANSACTIONS ON IMAGE PROCESSING (2010-) and the IEEE TRANSACTIONS ON CIRCUITS AND SYSTEMS FOR VIDEO TECHNOLOGY (2006-). He is an elected member of the IEEE Image and Multidimensional Signal Processing Technical Committee (2007-), the IEEE Visual Signal Processing and Communications Technical Committee (2006-), and the IEEE Multimedia Systems and Applications Technical Committee (2005-). He has served as Vice-Chair of the IEEE Multimedia Communications Technical Committee (2004-2006) and as a member of the IEEE Multimedia Signal Processing Technical Committee (2004-2007). He received the Swiss NSF Professorship Award in 2003, the IBM Faculty Award in 2005 and the IBM Exploratory Stream Analytics Innovation Award in 2008.
\end{IEEEbiography}

\end{document}